\documentclass[conference]{IEEEtran}
\IEEEoverridecommandlockouts
\usepackage{cite}
\usepackage{amsmath,amssymb,amsfonts}
\usepackage{algorithm}
\usepackage{algorithmic}
\usepackage{graphicx}
\usepackage{subfigure}
\usepackage{ragged2e} 
\usepackage{multirow}
\usepackage{array}
\usepackage[normalem]{ulem}
\usepackage{textcomp}
\usepackage[hidelinks]{hyperref}
\usepackage{xcolor}
\usepackage{makecell}
\hypersetup{
    colorlinks=true,
    linkcolor=blue,
    filecolor=blue,
    urlcolor=blue,
    citecolor=blue
}
\begin{document}

\title{Prompting-in-a-Series: Psychology-Informed Contents and Embeddings for Personality Recognition with Decoder-Only Models 
\thanks{This research was supported by Universiti Tunku Abdul Rahman Research Fund (IPSR/RMC/UTARRF/2021-C1/K03)}
}

\newcommand\correspondingauthor{\thanks{Corresponding author.}}

\author{\IEEEauthorblockN{Jing Jie Tan\IEEEauthorrefmark{1},
Ban-Hoe Kwan\IEEEauthorrefmark{1}, 
Danny Wee-Kiat Ng\IEEEauthorrefmark{1},
Yan-Chai Hum\IEEEauthorrefmark{1}},
Anissa Mokraoui\IEEEauthorrefmark{2},
Shih-Yu Lo\IEEEauthorrefmark{3}.

\IEEEauthorblockA{\IEEEauthorrefmark{1}Department of Mechatronics and Biomedical Engineering, Lee Kong Chian Faculty of Engineering and Science,\\ Universiti Tunku Abdul Rahman, Malaysia\\
}
\IEEEauthorblockA{\IEEEauthorrefmark{2}Laboratoire de traitement et transport de l'information, Université Sorbonne Paris Nord, France
}
\IEEEauthorblockA{\IEEEauthorrefmark{3}Institute of Communication Studies, National Yang Ming Chiao Tung University, Taiwan
}
Email: tanjingjie@1utar.my, \{kwanbh,ngwk,humyc\}@utar.edu.my, anissa.mokraoui@univ-paris13.fr, shihyulo@nycu.edu.tw 
}

\maketitle

\begin{abstract}
Large Language Models (LLMs) have demonstrated remarkable capabilities across various natural language processing tasks. This research introduces a novel "Prompting-in-a-Series" algorithm, termed PICEPR (Psychology-Informed Contents Embeddings for Personality Recognition), featuring two pipelines: (a) Contents and (b) Embeddings. The approach demonstrates how a modularised decoder-only LLM can summarize or generate content, which can aid in classifying or enhancing personality recognition functions as a personality feature extractor and a generator for personality-rich content. We conducted various experiments to provide evidence to justify the rationale behind the PICEPR algorithm. Meanwhile, we also explored closed-source models such as \textit{gpt4o} from OpenAI and \textit{gemini} from Google, along with open-source models like \textit{mistral} from Mistral AI, to compare the quality of the generated content. The PICEPR algorithm has achieved a new state-of-the-art performance for personality recognition by 5-15\% improvement. The work repository and models' weight can be found at \href{https://research.jingjietan.com/?q=PICEPR}{https://research.jingjietan.com/?q=PICEPR}.

\end{abstract}


\section{Introduction}
In recent years, Large Language Models (LLMs) have gained significant popularity for chatbot applications \cite{dam2024completesurveyllmbasedai}, becoming the backbone of many modern AI applications, including finance \cite{zhao2024revolutionizingfinancellmsoverview} and the arts \cite{choi2024proxonaleveragingllmdrivenpersonas}. Their ability to understand and generate human-like text has revolutionized various fields, from chatbots to content creation. However, the importance of explainability remains crucial \cite{cambria2024xaimeetsllmssurvey}, motivating us to explore the use of LLMs for personality recognition. Personality recognition plays a key role in enhancing AI's explainability \cite{NEUGEBAUER20211003,takami2023, Tan2025}, particularly in applications such as recommendation systems, virtual assistants, and human-computer interaction. By integrating personality recognition, AI systems can provide more personalized, context-aware, and user-aligned responses, making them not only more efficient but also more trustworthy and relatable.

\section{Literature Review}
This section explores three components: personality theories, Large Language Models (LLMs), and personality recognition.

\subsection{Personality Theories}
Personality serves as a psychological framework for understanding and explaining human behavior. Over time, various theories have been developed to classify and describe different aspects of personality, each proposing unique dimensions or traits. At its core, facets represent the foundational, measurable traits that shape personality, reflecting specific patterns of thought, feeling, and behavior. These facets combine to form broader dimensions, offering a nuanced understanding of individual differences \cite{Irwing2023}. Popular models of personality include the Myers-Briggs Type Indicator (MBTI), which categorizes personality into four dichotomous dimensions \cite{Schmitt2007,Obschonka2013}, and the Big-5 model, also known as OCEAN, which identifies five fundamental personality traits: Openness, Conscientiousness, Extraversion, Agreeableness, and Neuroticism \cite{Goldberg1993}. Personality serves as a vital tool for explaining human behavior, with applications in predicting buying intent, shaping advertisement content, and may other domains.

\subsection{Large Language Model}
Since Bidirectional Encoder Representations from Transformers (BERT) was introduced with a transformer-based architecture, it laid the foundation for the development of Large Language Models (LLMs), which leverage the transformer’s ability to capture long-range dependencies and contextual relationships in text \cite{koroteev2021bert}. These LLMs, such as Generative Pretrained Transformer (GPT) and Text-to-Text Transfer Transformer (T5), have significantly advanced natural language understanding and generation by scaling up model size and pre-training on vast amounts of data, enabling impressive performance across a wide range of tasks with minimal fine-tuning\cite{gpt4o_2024,team2024gemini}. Approaches like prompting leverage the pretrained capabilities of LLMs without requiring any fine-tuning, guiding their behavior through carefully crafted input instructions. However, fine-tuning LLMs can be employed to adapt the model more specifically to a task.

\subsubsection{Prompting}
Prompting eliminates the need for retraining and introduces a novel method that leverages unsupervised learning \cite{sahoo2024systematicsurveypromptengineering}. This approach enables models to handle new tasks without extensive training, including zero-shot \cite{radford2019language}, one-shot, and few-shot prompting \cite{brown2020language}.

\begin{itemize}
    \item \textbf{Zero-shot prompting}: The model is given a prompt to perform a task without any specific examples, relying solely on general knowledge from its pretraining.
    \item \textbf{One-shot prompting}: The model receives one example of the desired task to guide its response, enabling a minimal amount of task-specific context.
    \item \textbf{Few-shot prompting}: The model is given a small number of examples to illustrate the task, allowing it to better align its output with task-specific patterns.
\end{itemize}

Given the focus of this study on personality recognition, we further explore the application of logical and reasoned approaches to understand how personality traits influence thought patterns. Researchers have proposed various prompting techniques to enhance the reasoning capabilities of LLMs. 
\begin{itemize}
    \item \textbf{Chain-of-Thought (CoT) Prompting:} Facilitates step-by-step reasoning, enabling models to articulate a logical process before reaching a conclusion, which is crucial for comprehending intricate personality-related patterns \cite{wei2022chain}.
    
    \item \textbf{Automatic Chain-of-Thought (Auto-CoT):} Builds on CoT by automatically generating reasoning chains without user intervention, eliminating the need for manual input and expanding the range of tasks the model can effectively handle \cite{zhang2022automatic}.
    
    \item \textbf{Logical Thought (LoT) Prompting:} Incorporates a self-improvement step using \textit{Reductio ad Absurdum} to systematically verify and refine the reasoning process, improving the accuracy and reliability of logical reasoning \cite{zhao2023enhancing}.
    
    \item \textbf{Chain-of-Symbol (CoS) Prompting:} Uses symbolic language for complex representations like relational and spatial reasoning, enhancing clarity and minimizing ambiguity \cite{hu2023chain}.
    
    \item \textbf{Tree-of-Thoughts (ToT) Prompting:} Introduces hierarchical exploration of ideas, allowing the model to evaluate and refine thoughts iteratively, aligning with structured personality assessments \cite{yao2024tree}.
    
    \item \textbf{Chain-of-Table Framework:} Leverages tabular data in reasoning chains for table understanding tasks, achieving state-of-the-art performance on multiple benchmarks \cite{wang2024chain}.
\end{itemize}

\subsubsection{Fine-Tuning}
Fine-tuning is a process used to adapt large pre-trained language models to specific tasks or domains by updating their weights with labeled task-specific data through supervised learning  \cite{Patil2024}. Various approaches have been studied to improve generalisation for unseen tasks. 

\textbf{Parameter-Efficient Fine-Tuning (PEFT)} is applied as traditional fine-tuning methods require updating all model parameters, which can be resource-intensive.  \begin{itemize}
    \item \textbf{Low-Rank Adaptation (LoRA):} Adjusts only a subset of parameters, significantly reducing computational demands while maintaining performance by introducing trainable low-rank matrices into each layer of the pre-trained model during fine-tuning, enabling efficient adaptation to downstream tasks \cite{han2024parameterefficientfinetuninglargemodels}.
    
    \item \textbf{DLoRA (Distributed LoRA):} Enables scalable parameter-efficient fine-tuning by distributing the process between cloud servers and user devices, addressing privacy concerns associated with sharing sensitive data in public environments \cite{wu2024dlora}.
    
    \item \textbf{QLoRA (Quantized LoRA):} Reduces memory usage by backpropagating gradients through a frozen, 4-bit quantized pre-trained language model into Low-Rank Adapters, allowing the fine-tuning of large models on single GPUs without compromising performance \cite{NEURIPS2023_1feb8787}.

\end{itemize}
    
Moving on, \textbf{Representation Fine-Tuning (ReFT)} is proposed to avoid updating the model weights. Instead, it modifies the hidden representations (intermediate outputs) generated by a frozen pre-trained model during inference. Hence, ReFT applies task-specific adjustments directly to these representations, enabling lightweight adaptation to new tasks while preserving the pre-trained knowledge \cite{wu2024reftrepresentationfinetuninglanguage}.

\subsection{Personality Recognition}
As machine learning continues to advance, researchers have increasingly explored its application to personality recognition—a multidisciplinary field at the intersection of artificial intelligence and social science \cite{wen2024selfassessmentexhibitionrecognitionreview, ji2023chatgptgoodpersonalityrecognizer}. Early efforts employed contextual language embeddings; for example, one study introduced an input representation mechanism that processes text by converting it into vector embeddings based on linguostylistic features, including frequency, co-occurrence, and context (FCC) measures for classification \cite{Pavan}.

With the development of large language models (LLMs), researchers have further investigated their potential in personality inference. Notably, recent work has demonstrated that the ChatGPT-4 model can infer personality traits during interactive conversations \cite{peters2024largelanguagemodelsinfer}. Additionally, frameworks leveraging ChatGPT’s prompting capabilities have been proposed to assess human personality \cite{rao2023chatgptassesshumanpersonalities}. However, these advances raise a critical question: Can LLMs accurately infer a user’s personality based solely on their past social media posts or one-way written essays?

Zero-shot learning has been studied, and while simple prompts showed limited performance, incorporating knowledge about personality traits (e.g., definitions and frequent words) significantly improved results \cite{ganesan2023systematicevaluationgpt3zeroshot}. 
Researchers also proposed Text Augmentation for Embeddings (TAE), a LLM-based text augmentation framework that enhances the personality embeddings representation by leveraging LLM-generated semantic, sentiment, and linguistic analyses. Contrastive learning aligns these augmentations in embedding space to improve psycho-linguistic representation, and enriching personality labels further boosts performance \cite{Hu_He_Wang_Zhao_Shao_Nie_2024}. Various machine learning models have also been proposed to enhance the representation, including BERT+MLP \cite{Mehta2020} \cite{Amirhosseini2020}, Multi-Document Transformer (MD-T) \cite{Yang_Quan_Yang_Yu_2021}, Deep Graph Convolutional Network (DGCN)\cite{yang2023orders}, Psycholinguistic Tripartite Graph Network (TrigNet)\cite{yang-etal-2021-psycholinguistic} and etc.

Conversely, researchers have investigated the ChatGPT model and found that it demonstrates remarkable personality recognition capabilities, along with strong interpretability in its prediction results \cite{ji2023chatgptgoodpersonalityrecognizer}. Lastly, studies have explored leveraging emotion knowledge to determine personality by combining results derived from prompting \cite{li2024eerpdleveragingemotionemotion}. Although this approach improves performance, it utilizes a decoder-only LLM to break down the training dataset, perform similarity measurement, and apply chain-of-thought (CoT) reasoning with additional samples. This approach combines several limitations: it requires pre-inferencing (though it demands relatively fewer resources than fine-tuning an encoder-only model) to build a reference library, and it performs slower than encoder-only models during classification. Therefore, future improvements could focus on reducing these drawbacks, while bringing us some fundamental insight into modularisation.

\subsection{Research Question}
Hence, this research aims to address the following key research questions (RQ):
\begin{enumerate}
    \item Do decoder-only LLMs benefit from a modularisation approach to extend CoT reasoning in personality classification?
    \item What is the impact of different LLM backbones on performance and reliability in personality classification tasks?
    \item Can decoder-only LLMs be leveraged more effectively—and in comparison to encoder-only LLMs—to provide psychological insights for personality recognition?
\end{enumerate}

\section{Methodologies}
This research consists of 4 steps in a pipeline: i) Dataset
ii) Algorithms; iii) Experiment Design; and iv) Evaluation. 
LLM research for personality recognition has not been extensively explored and lacks sufficient baselines. Therefore, we will compare it with a transformer model, LSTM, and classical NLP approaches while also establishing a baseline, such as direct classification using LLM.

\subsection{Dataset Preparation}
This research utilized the Essays dataset that utilised the Big-5 personality model \cite{Pennebaker1999} \cite{Mairesse2007} and the Kaggle dataset that utilized the MBTI personality model \cite{mitchell2017}. The Table \ref{tab:dataset_stats} below summarizes the datasets as well as the binary personality dimensions in each theory. The datasets were split into training, validation, and test sets according to Tan's train-validation-evaluation split method \cite{essays-dataset, kaggle-dataset} across this research to ensure fair comparison and to verify the effectiveness of our model. Additionally, it is important to note that the Essays dataset was collected in a formal setting, resulting in texts with a more formal style \cite{Mairesse2007}. This characteristic makes the Essays dataset inherently more challenging, thereby limiting the potential for higher accuracy. Moreover, the Kaggle dataset exhibits significant label imbalance, with ratios exceeding 6:4. An asterisk (*) next to a dimension indicates such imbalance between the 0 and 1 labels in the respective dataset. This imbalance often leads to reduced accuracy. The table below summarizes the datasets as well as the binary personality dimensions in each theory.

\begin{table}[htbp]
\scriptsize
  \centering
  \caption{Dataset Statistics and Personality Dimensions}
  \begin{tabular}{|l|l|l|}
    \hline
    \textbf{Dataset} & \textbf{Essays} & \textbf{Kaggle} \\
    \hline
    \makecell[l]{Number of Samples \\(Train, Validation, Test)} & \makecell[l]{2467 \\(1578, 395, 494)} & \makecell[l]{8675 \\(5552, 1388, 1735)} \\
    \hline
    \multirow{5}{*}{Personality Dimensions} 
    & Openness (O) & Sensing/ Intuition (S/N)* \\
    & Conscientiousness (C) &Perceiving/ Judging (P/J)*  \\
    & Extraversion (E) &  Intro-/ Extra-version (I/E)*  \\
    & Agreeableness (A) & Thinking/ Feeling (T/F)\\
    & Neuroticism (N) & - \\
    \hline
  \end{tabular}
  \label{tab:dataset_stats}
\end{table}

In this research, we minimize the dataset's text preprocessing before inputting it into the Large Language Model (LLM) to fully leverage the inherent capabilities of pre-trained models in processing raw or lightly processed text. LLMs are designed to learn universal language representations from extensive datasets, allowing them to handle diverse linguistic structures and nuances without extensive feature engineering \cite{Kumar2024}. This approach enables us to evaluate how LLMs perform on unstructured, conversational data directly sourced from social media. Meanwhile, we also treat it as multilabel classification instead of 4-5 binary classification tasks to avoid repeat training for each dimension, thereby reducing training time and computational resources.

\subsection{Algorithms}
\begin{figure*}
    \centering
    \includegraphics[width=1\textwidth]{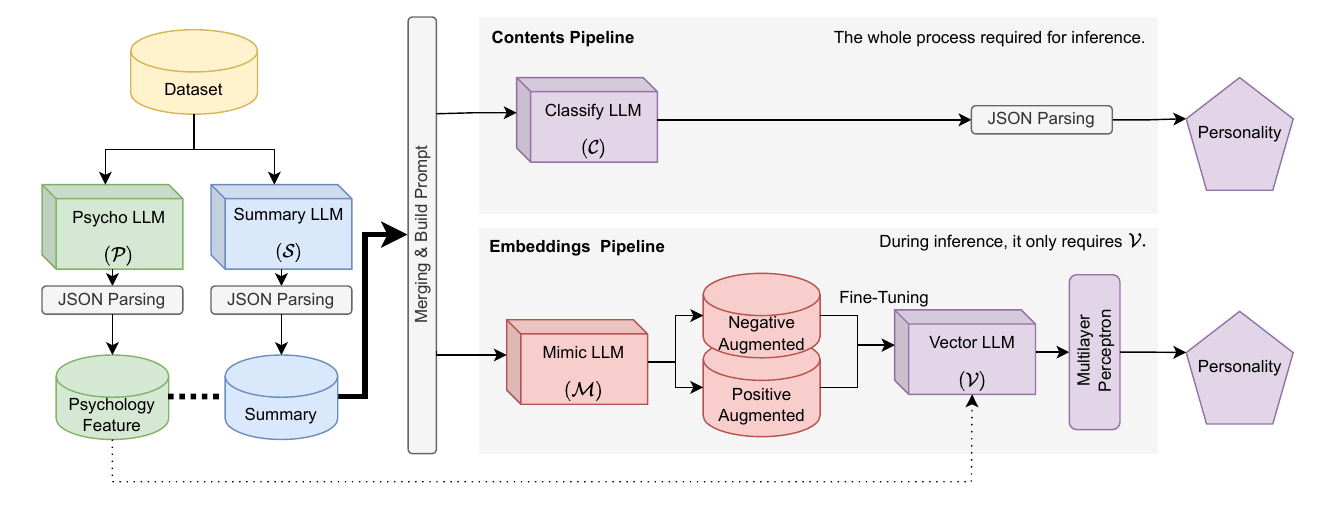}
    \caption{Algorithmic Workflow of the PICEPR Framework (Psychology-Informed Contents and Embeddings for Personality Recognition): The diagram presents a structured overview of the dual-pipeline architecture (a) Contents Pipeline and (b) Embeddings Pipeline underlying the PICEPR algorithm. Rectangular nodes signify the four integral components of Large Language Models (LLMs), while the yellow cylinder represents the initial, unprocessed dataset. Colored cylinders illustrate various intermediate datasets generated throughout the process via LLM interactions. Note that the dotted line represents optional content or embeddings included during empirical testing, which excluded based on ablation feedback.}
    \label{fig:PICEPR-visualisation}
\end{figure*}

To address the research question, we proposed with the Psychology-Informed Contents Embeddings for Personality Recognition (PICEPR) algorithm, illustrated in Fig.~\ref{fig:PICEPR-visualisation}. Our contributions are as follows:

\begin{itemize}
    \item We \textbf{analyse whether Prompting-in-a-Series framework, a modularisation approach can further extend Chain-of-Thought (CoT) reasoning} for personality classification using decoder-only models, or generate high-quality datasets for fine-tuning encoder-only models.
    
    \item We \textbf{study performance of various open- and closed-source large language models (LLMs) on personality tasks}, which psychology-related task have not been extensively studied in the current AI era.
    
    \item We \textbf{analyze whether decoder-only LLMs are capable of eliminating the need for task-specific fine-tuning or assistance from referencing libraries}, in comparison to training encoder-only models for personality tasks.
\end{itemize}

The PICEPR algorithm is composed of five distinct components, each designed to harness the unique capabilities of a specific Large Language Model (LLM): \textbf{Summary LLM ($\mathcal{S}$)}, \textbf{Mimic LLM ($\mathcal{M}$)}, \textbf{Psycho LLM ($\mathcal{P}$)}, \textbf{Classify LLM ($\mathcal{C}$)}, and \textbf{Vector LLM ($\mathcal{V}$)}. The algorithm operates via two distinct pipelines: \textit{Contents} and \textit{Embeddings}. 

\begin{itemize}
    \item \textbf{Contents Pipeline:} This pipeline involves the components $\mathcal{S}$, $\mathcal{P}$, and \textbf{$\mathcal{C}$}. It relies exclusively on the prompting step to perform its tasks. The advantage is that it does not require fine-tuning but can classify personality in a short, reasonable inference time.
    
    \item \textbf{Embeddings Pipeline:} This pipeline incorporates the components $\mathcal{S}$, $\mathcal{P}$, \textbf{$\mathcal{M}$} and \textbf{$\mathcal{V}$}. Unlike the Contents Pipeline, it can migrate the decoder-only LLM knowledge through data augmentation and fine-tune an encoder-only model that provides personality classification at a relatively fast speed and low computational cost.
\end{itemize}

From the literature, we learn that Chain-of-Thought (CoT) prompting provides better performance compared to directly outputting the required label. Therefore, we ensure that analyses were generated as a response before outputting the final label. We explicitly separate the \textit{system} role and the \textit{user} role in our prompting (whenever possible) to streamline the task. The system role defines task, as well as the input format and output structure, while the user role supplies the input required to generate the output, which is structured according to the format specified by the system role (if applicable).

We also focus on using structured output to facilitate the abstraction of the intended data. However, not all models support structured output (JSON schema), which can result in minor formatting errors (e.g., \texttt{"\textbackslash n"} or extra symbols)
. We first attempt to use the structured output engine \cite{willard2023efficientguidedgenerationlarge} to output the JSON format according to the system prompt (if not applicable, the instruction is placed in the user prompt). To ensure a fair comparison and enough valid data for experiment, we employ a JSON repair tool to correct any issues \cite{Baccianella_JSON_Repair_-_2024}. If the repair tool does not resolve the problem, we mark it as an error, exclude the corresponding row, and continue with the experiment.

\subsubsection{Summary LLM ($\mathcal{S}$)}
\begin{figure}[ht]
    \centering
    \fbox{\includegraphics[width=0.48\textwidth]{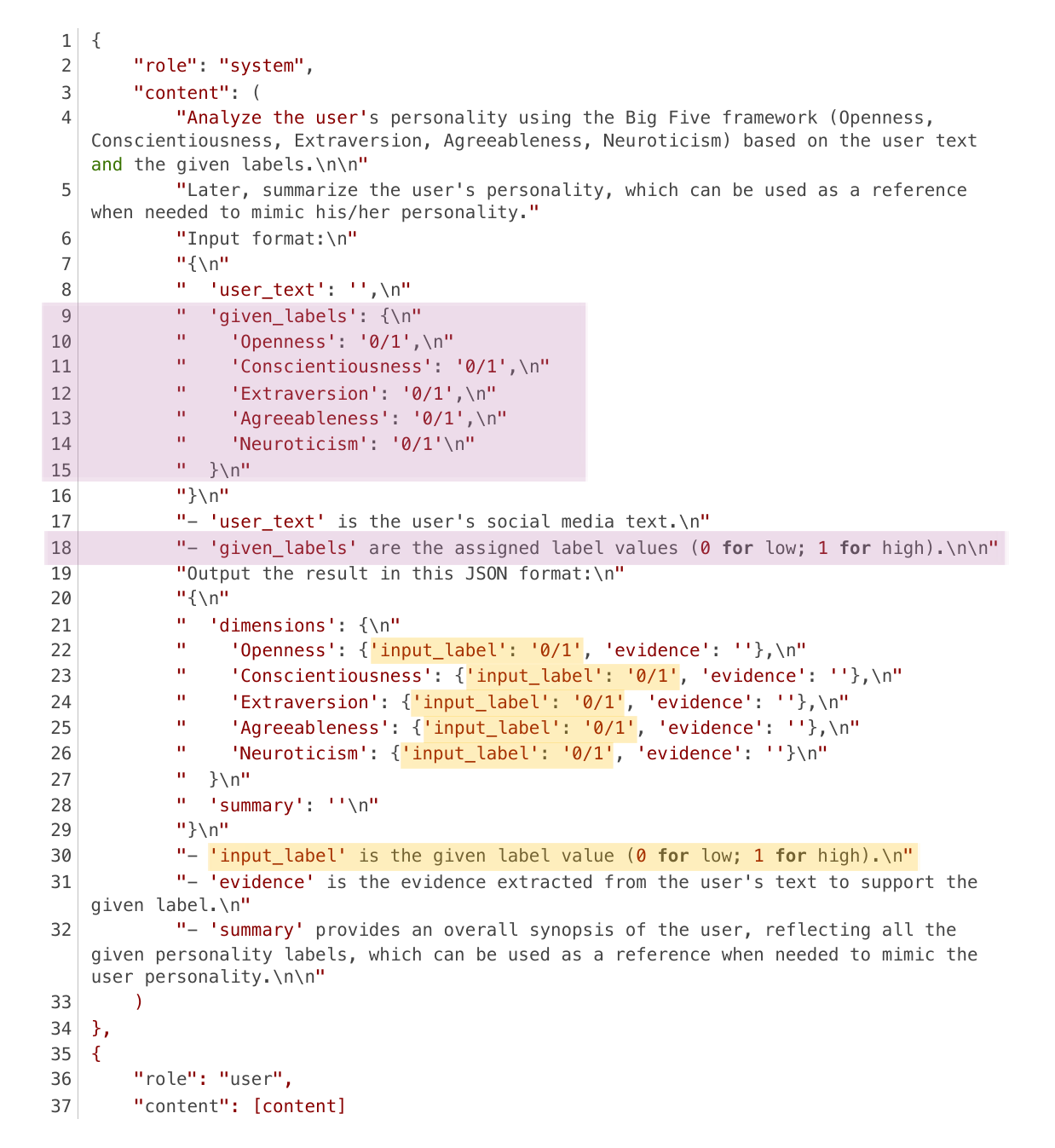}}
    \caption{The Chain of Thought (CoT) Prompt for PICEPR's \textbf{Summary LLM ($\mathcal{S}$)} is designed to provide a synopsis of user personality based on the given \textit{user\_text} from the dataset.The highlighted content (lines 9–15, 18, 22–27, and 30) will be removed during inferencing (test dataset). This applies to both the \textbf{Contents} and \textbf{Embeddings} pipeline.}
    \label{fig:SummaryLLMPrompt}
\end{figure}

This component is designed to provide a concise overview of the user’s personality traits. As illustrated in Fig. \ref{fig:SummaryLLMPrompt}, Chain of Thought (CoT) prompting is utilized to ensure that the summary accurately reflects the user's personality as identified from the text. The inclusion of ``given" labels highlights the predefined personality traits that are being analyzed, effectively guiding the LLM throughout the processing. To enhance coherence, the output incorporates the input labels along with evidence extracted from the text, ensuring consistency between the identified traits and their supporting information. Ultimately, the summary integrates these dimensions and evidence into a cohesive representation, offering a dependable reference for mimicking the user's personality traits. However, this is only for training purposes; for experiments involving the .test set, the labels will not be provided to ensure reliable recognition results.

\subsubsection{Psycho LLM ($\mathcal{P}$)}
\begin{figure}[ht]
    \centering
    \fbox{\includegraphics[width=0.48\textwidth]{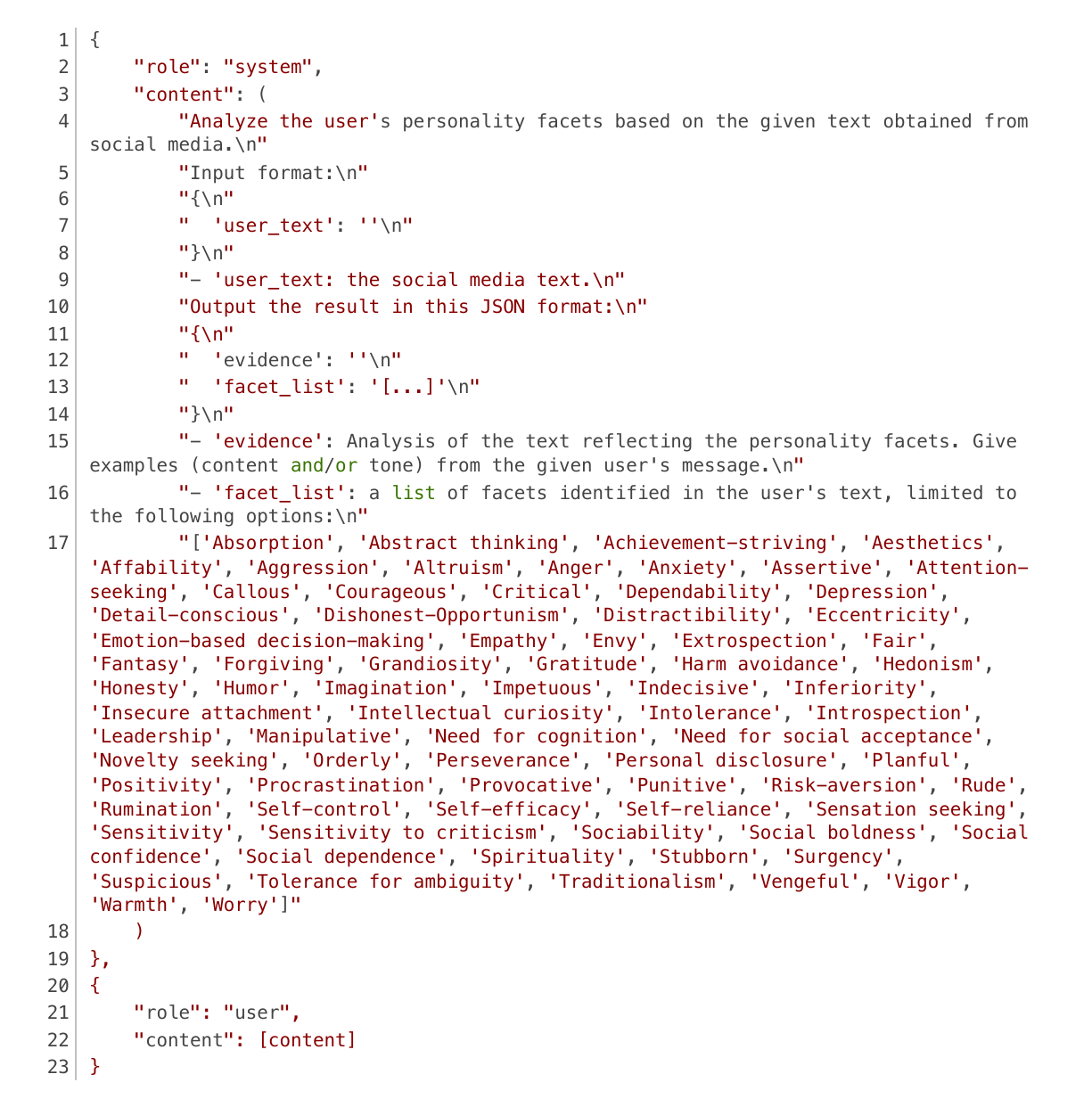}}
    \caption{The Chain of Thought (CoT) Prompt for PICEPR's \textbf{Psycho LLM ($\mathcal{P}$)} to provide a \textit{facet\_lists} according to 77 personality facets \cite{Irwing2023} using the given \textit{user\_text} from the dataset. This applies to both the \textbf{Contents} and \textbf{Embeddings} pipeline. }
    \label{fig:PsychoLLMPrompt}
\end{figure}

Psycho LLM is designed to analyze user personality facets. This LLM aims to evaluate 77 personality facets based on the latest personality research, effectively capturing the nuances of user personality \cite{Irwing2023}. For each facet, the LLM provides a binary score of 0 or 1, which labels the presence or absence of the trait. The results are output in JSON format, making them readily usable as features for further applications. Fig. \ref{fig:PsychoLLMPrompt} illustrates the designed prompt.

\subsubsection{Classify LLM ($\mathcal{C}$)}
Classify LLM serves as a core component for producing personality labels. Our experiments involve regular Chain-of-Thought (CoT) classification, fine-tuning, and the PICEPR algorithm to evaluate whether the PICEPR algorithm enhances performance and provides deeper insights into behavior inferred from text on social networks.

\begin{figure}[ht]
    \centering
    \fbox{\includegraphics[width=0.48\textwidth]{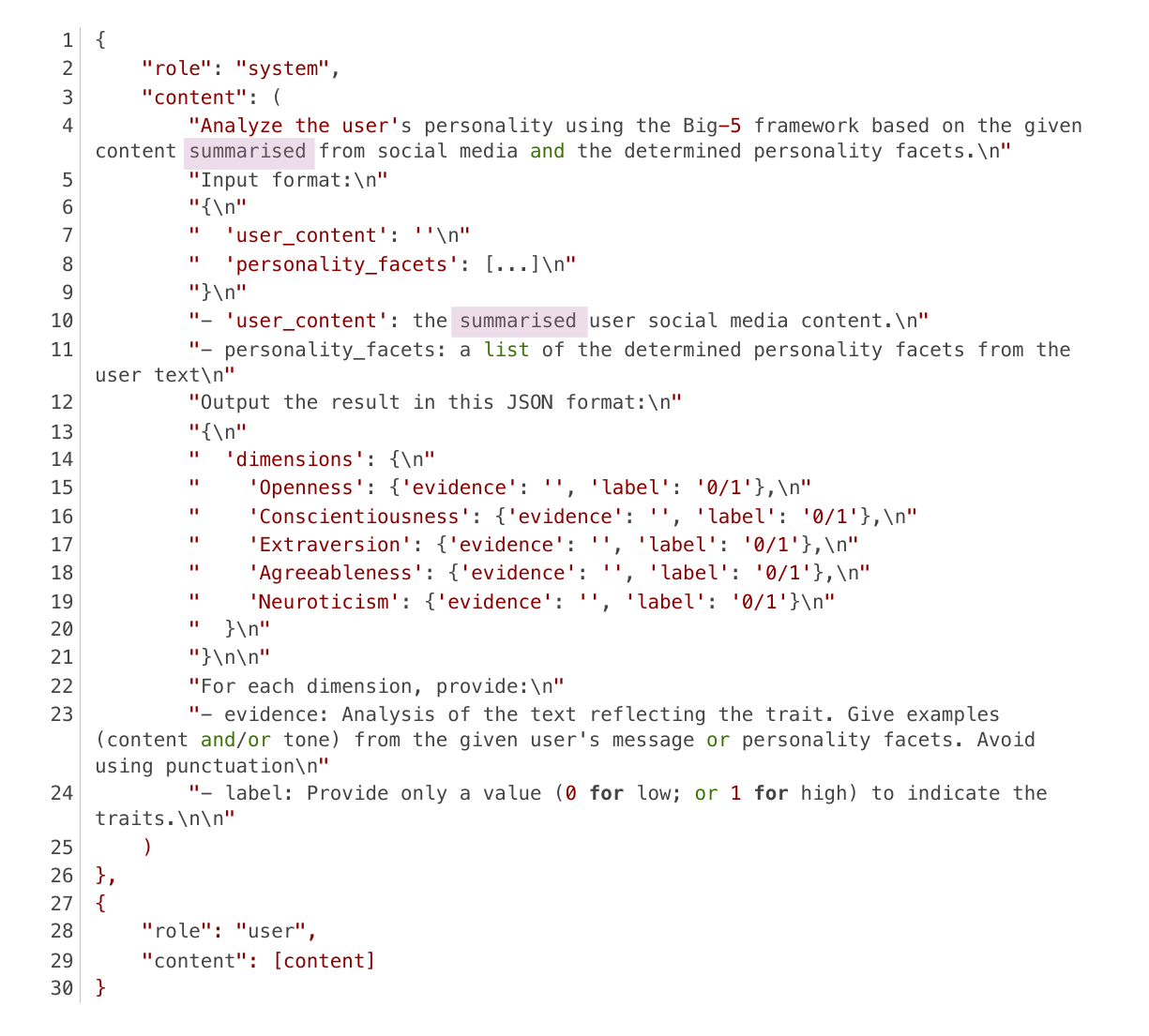}}
\caption{The Chain-of-Thought (CoT) prompt for PICEPR's \textbf{Contents} Pipeline's \textbf{Classify LLM ($\mathcal{C}$)} is used to generate personality labels. The \textit{user\_Contents} is derived from $\mathcal{S}$, and the \textit{personality\_facets} are obtained from $\mathcal{P}$. The red-highlighted parts from the system prompt are excluded. Additionally, the user content is replaced with the original content, and the personality facets are removed during the baseline setup for standard CoT classification.
}
    \label{fig:ClassifyLLMPrompt}
\end{figure}

In the PICEPR algorithm, the model processes content generated by the Summary LLM ($\mathcal{S}$) alongside the personality facet outputs from the Psycho LLM ($\mathcal{P}$). Fig. ~\ref{fig:ClassifyLLMPrompt} illustrates the prompt employed by the Classify LLM ($\mathcal{C}$) to structure a JSON output for personality types. A similar prompt is also used to establish a baseline for regular CoT classification.

For fine-tuning, we compare the PICEPR method with a fine-tuned Classify LLM (\(\mathcal{C}\)). The LLM is a decoder-only model, and no existing dataset is available for this specific task. To address this, we utilize the Summary LLM (\(\mathcal{S}\)) to emulate the chat completion output (\texttt{"role"="assistant"}) by combining the evidence (analysis) component with the corresponding label. This approach allows us to construct a training set containing targeted outputs derived from users' social media text and our system prompt. Eq.\ref{eq:cross_entropy} shows the cross-entropy loss used for fine-tuning.

\begin{equation}
\label{eq:cross_entropy}
\mathcal{L}_{\text{CE}} = - \sum_{t=1}^{T} y_{t} \cdot \log\left( p_{t} \right)
\end{equation}

Where:
\begin{itemize}
    \item \(T\): Number of tokens.
    \item \(y_{t}\): True distribution for the \(t\)-th token.
    \item \(p_{t}\): Predicted probability for the \(t\)-th generated token.
\end{itemize}

 However, due to the computation resources limitation, we employ Quantized Low-Rank Adaptation (QLoRA) to accelerate the fine-tuning process. The QLoRA weight update is defined as shown in Eq. \ref{eq:lora}.

\begin{equation}
    \label{eq:lora}
    W' = W + \Delta W =  W + A \cdot B
\end{equation}

Where:
\begin{itemize}
    \item $W \in \mathbb{R}^{d \times k}$: Pre-trained weight matrix.
    \item $A \in \mathbb{R}^{d \times r}$: Low-rank matrix 1.
    \item $B \in \mathbb{R}^{r \times k}$: Low-rank matrix 2.
    \item $d$: Input dimension of the original weight matrix.
    \item $k$: Output dimension of the original weight matrix.
    \item $r$: Rank of the low-rank decomposition, $r \ll \min(d, k)$.
\end{itemize}

\subsubsection{Mimic LLM ($\mathcal{M}$)}
\begin{figure}[ht]
    \centering
    \fbox{\includegraphics[width=0.48\textwidth]{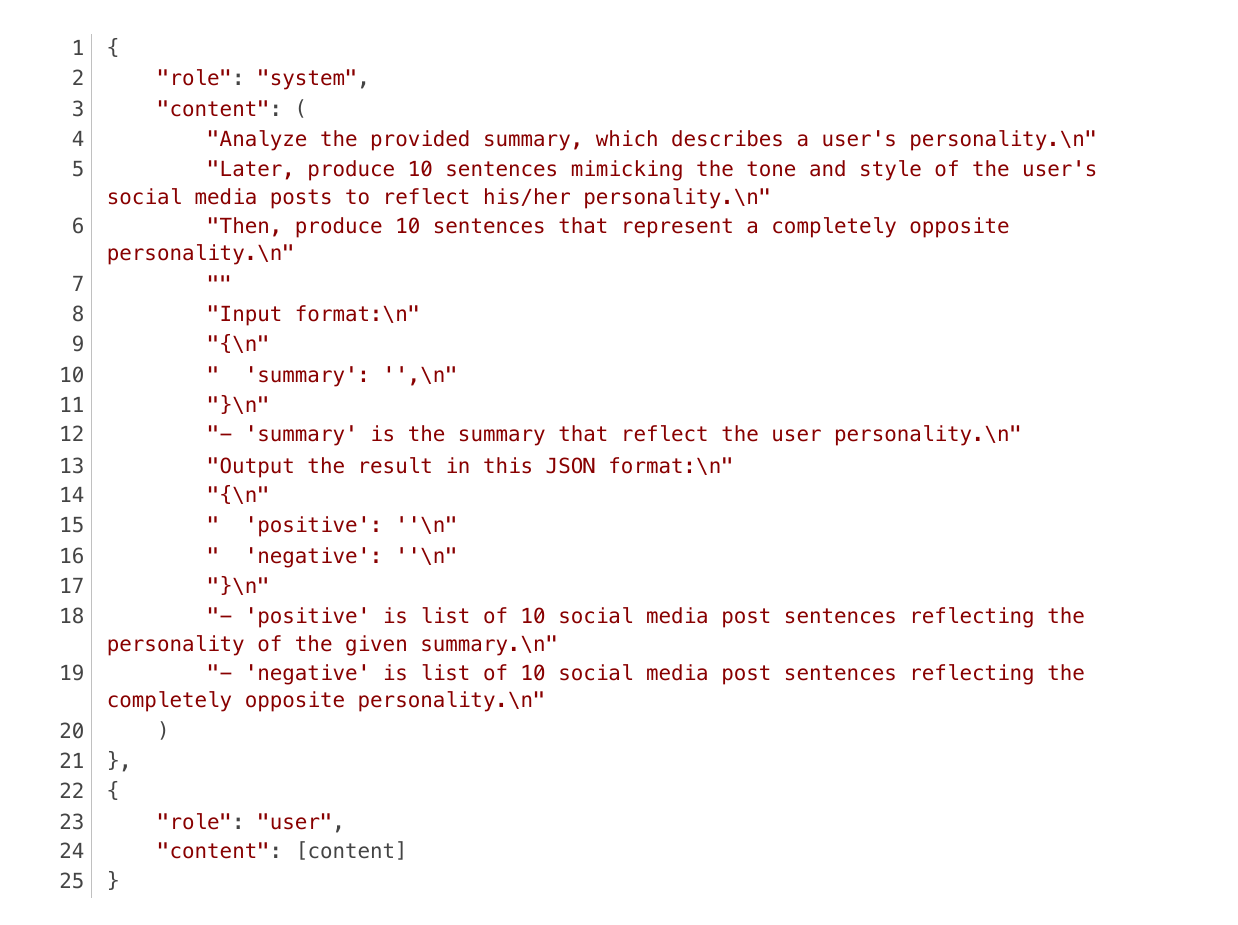}}
    \caption{The Chain-of-Thought (CoT) prompt for PICEPR's \textbf{Embeddings} Pipeline's \textbf{Mimic LLM ($\mathcal{M}$)} is used to generate augmented positive and negative social media content. The \textit{summary} is derived from $\mathcal{S}$, and personality facets from $\mathcal{P}$ may be merged into the input.}

    \label{fig:MimicLLMPrompt}
\end{figure}

As shown in Fig. \ref{fig:MimicLLMPrompt}, this component generates an augmented dataset by leveraging the summarized content and corresponding labels from the Summary LLM ($\mathcal{S}$). The Mimic LLM further synthesizes social media-style text that mirrors the personality traits indicated by the labels. This approach enriches the dataset by incorporating realistic, personality-reflective examples, ensuring that the augmented data is contextually relevant and diverse. Additionally, the process involves prompting the model to produce a negative set—examples that do not align with the targeted personality traits—enabling more effective fine-tuning. Together, the augmented dataset contribute to improving the robustness and accuracy of downstream personality recognition tasks by enhancing the model's ability to generalize across various textual nuances.

\subsubsection{Vector LLM ($\mathcal{V}$)}
This is also a core component for the experiment, but this differs from the aforementioned Mimic LLM ($\mathcal{M}$). The Vector LLM (\(\mathcal{V}\)) processes user text and converts it into embeddings of length \(n\), where $n$ is the length of the embedding vector determined by the specific type and architecture of the LLM. These embeddings serve as dense numerical representations of the processed content, capturing meaningful patterns and relationships.

Since this model outputs only embeddings, a multilayer perceptron (MLP) is employed to classify personality into the final categories. The MLP’s input layer has size $r + 77$, where $r$ is the length of the representation embedding vector, which varies depending on the choice of $\mathcal{V}$, and 77 corresponds to the additional personality facet outputs from $\mathcal{P}$ \cite{Irwing2023}. Similar to $\mathcal{C}$, this setup allows for studying whether the PICEPR algorithm can enhance performance and provide better insights into the LLM’s behavior.

As there are 4 or 5 labels depending on the personality theory applied, this is treated as a multitask classification. The loss function, \(\mathcal{L}_{\text{MBCE}}\), for training is multi-binary cross-entropy, as shown in Eq. \ref{eq:mbce}, combined with a sigmoid activation function, defined in Eq. \ref{eq:sigmoid}.

\begin{equation}
\label{eq:mbce}
\mathcal{L}_{\text{MBCE}}\left(y, \hat{y} \right) = \sum_{c=1}^C -\left[ y_{c} \log \hat{y}_{c} + (1 - y_{c}) \log (1 - \hat{y}_{c}) \right]
\end{equation}

\begin{equation}
\label{eq:sigmoid}
\hat{y}_{c} = \frac{1}{1 + e^{-z_{c}}}
\end{equation}

Where:
\begin{itemize}
    \item \( C \): Number of classes (4 for MBTI  \& 5 for Big-5).
    \item \( y_{c} \): True label (1 or 0) for class \(c\).
    \item \( \hat{y}_{c} \): Predicted probability for class \(c\).
    \item \( z_{c} \): Logit output before applying the sigmoid function.
\end{itemize}

To effectively address the issue of class imbalance, the focal loss is utilized, as defined in Eq. \ref{eq:focal}, where $\alpha$ and $\gamma$ are the hyper-parameters. This loss function modifies the regular cross-entropy loss by incorporating a modulating factor that emphasizes harder-to-classify examples while down-weighting the contribution of well-classified ones. 

\begin{equation}
\label{eq:focal}
L_{\text{FBCE}}\left(y, \hat{y} \right) = \alpha \cdot \left(1 - e^{-L_{\text{MBCE}}\left(y, \hat{y} \right)}\right)^\gamma \cdot L_{\text{MBCE}}\left(y, \hat{y} \right)
\end{equation}

We adapt the contrastive loss $\mathcal{L}_{\text{contrastive}}$ (Eq.~\ref{eq:contrastiveloss}) as the loss function during fine-tuning. The objective is to maximize the similarity value of positive pairs (same personality) while minimizing it for negative pairs.

\begin{equation}\label{eq:contrastiveloss}
\mathcal{L}_{\text{contrastive}} = yD^2 + (1 - y) \max(0, m - D)^2,
\end{equation}

where:
\begin{itemize}
    \item $D$: Euclidean distance of vectors $a$ and $b$, $ \| d^{(a)} - d^{(b)} \|$.
    \item $y$: Binary label indicating pair type, $y \in \{0, 1\}$.
    \item $m$: Minimum threshold distance for negative pairs.
\end{itemize}

\subsection{Experiment Setting}
We designed a series of experiments to establish performance baselines for both pipelines. As listed in Table~\ref{tab:LLM-list}, we utilized various LLMs as backbones. Additionally, we experimented with combinations of encoder and decoder architectures in the Embeddings Pipeline, guided by empirical performance results.

For the \textbf{Contents} pipeline, we employed the following configurations: (a) Chain-of-Thought (CoT) prompting for standard classification ($\mathcal{C}^{\text{R}}$), (b) fine-tuning the LLM ($\mathcal{C}^{\text{RT}}$), (c) our proposed PICEPR algorithm ($\mathcal{C}^{\text{PR}}$), and (d) its two-shot prompting variant ($\mathcal{C}^{\text{PR2}}$). On the other hand, for the \textbf{Embeddings} pipeline, we explored: (a) standard LLM-based classification ($\mathcal{V}^{\text{R}}$), (b) fine-tuning the encoder model ($\mathcal{V}^{\text{RT}}$), (c) adopting an augmented dataset generated by a decoder LLM ($\mathcal{M}'$) to fine-tune the encoder model ($\mathcal{V}^{\text{AT}}$), and (d) the PICEPR algorithm ($\mathcal{V}^{\text{PR}}$). 

\begin{table}[h!]
\centering
\scriptsize
\caption{List of LLMs used in the research, along with their respective symbols, versions, structure, and types.}
\label{tab:LLM-list}
\begin{tabular}{|l|l|l|l|}\hline

\textbf{Abbreviation} & \textbf{Model Version}   & \textbf{Structure} & \textbf{Type} \\ \hline
\textit{gpt4o}  \cite{gpt4o_2024}  & gpt-4o-2024-08-06  & Schema& Completions \\ \hline
\textit{gpt3.5} \cite{gpt3.5_turbo_2024}  & gpt-3.5-turbo-0125 & Mode & Completions \\ \hline
\textit{gemini} \cite{team2024gemini}  & gemini-1.5-flash & Schema & Completions \\ \hline
\textit{llama} \cite{llama3.2_2024}   & Meta-Llama-3.1-8B-Instruct  & Text & Completions \\ \hline
\textit{mistral} \cite{mistral_7b_2024} & Mistral-7B-Instruct-v0.3& Text & Completions \\ \hline
\textit{gptada} \cite{text_embedding_ada_002}  & text-embedding-ada-002  & Vector & Embeddings   \\ \hline
\textit{gemb} \cite{lee2024geckoversatiletextembeddings}  & text-embedding-004& Vector  & Embeddings   \\ \hline
\textit{mistemd} \cite{mistral_embed}  & mistral-embed-23.12& Vector & Embeddings   \\ \hline
\textit{minilm} \cite{wang2020minilmdeepselfattentiondistillation}  & paraphrase-MiniLM-L6-v2 & Vector & Embeddings   \\ \hline
\textit{mpnet}  \cite{all_mpnet_base_v2}  & all-mpnet-base-v2  & Vector  & Embeddings   \\ \hline
\textit{sbert} \cite{reimers-2019-sentence-bert} & sentence-bert-base-nli-mean-tokens   & Vector    & Embeddings \\ \hline
\end{tabular}

\begin{justify}
\footnotesize{
Note: The experiment uses 5 LLM components. To ensure consistency in notation throughout the paper, we adopt the following format: $\mathcal{LLM}_{\text{v}}$, where $\mathcal{LLM}$ denotes the corresponding LLM components, such as $\mathcal{S}$, $\mathcal{M}$, $\mathcal{P}$, or $\mathcal{C}$; and $\text{v}$ specifies the LLM model version. For instance, \textbf{$\mathcal{V}_{\text{minilm}}\{\mathcal{S},\mathcal{M},\mathcal{P}\}_{\text{gpt4o}} $} indicates the pipeline: Vector LLM ($\mathcal{V}$) using the \textit{minilm} model; Summary LLM , Mimic LLM ($\mathcal{M}$), and Psycho LLM ($\mathcal{P}$) using the \textit{gpt4o} model. In addition, not all LLM models can be all incorporated into the 5 components. For example, \textit{gptada} model generates only embeddings and does not produce text output, making it only can be utilized for Vector LLM ($\mathcal{V})$. In addition, even if the LLM documentation states that it supports JSON formatting, the output structure or values might still be incorrect. For model output structure, we indicate `Schema' and `Mode' (for cases where format failure is more likely) if it is explicitly supported according to the documentation, and `Text' when JSON output is not guaranteed but can still be achieved using the structured output engine \cite{willard2023efficientguidedgenerationlarge}.
}
\end{justify}
\end{table}

Last but not least, we also analyze the invalid generated outputs (which are the data that cannot be correctly parsed as valid classifications during the inference process), which impact the evaluation of large language models (LLMs) for personality classification.

\subsection{Evaluation}

To evaluate personality recognition performance, we employ standard metrics: Regular Accuracy (RA) (Eq.~\ref{eq:ra}) to assess the model accuracy, Balanced Accuracy (BA) (Eq.~\ref{eq:ba}) to measure model performance on imbalanced datasets (which should be nearly identical to RA for balanced datasets), and the F1 Score (Eq.~\ref{eq:f1}) to evaluate model bias. Here, $TP$, $FP$, $TN$, and $FN$ denote true positives, false positives, true negatives, and false negatives, respectively.

\begin{equation}
RA = \frac{TP + TN}{TP + TN + FP + FN}.
\label{eq:ra}
\end{equation}

\begin{equation}
BA = \frac{\left(\frac{TP}{TP + FN} + \frac{TN}{TN + FP}\right)}{2}.
\label{eq:ba}
\end{equation}

\begin{equation}
F1 = \frac{2 \cdot TP}{2 \cdot TP + FP + FN}.
\label{eq:f1}
\end{equation}

Additionally, we analyze the errors generated by each LLM, particularly focusing on instances where the model fails to produce a valid JSON output that can be parsed without the use of repair tools \cite{Baccianella_JSON_Repair_-_2024}, or where there are spelling errors in JSON keys or invalid values (e.g., a label of 0 or 1 but given output -1). The error rate (\(ER\)) is calculated as in Eq. \ref{eq:error-rate}.
\begin{equation}
ER = \frac{\text{json}_{\text{number of rows with errors}}}{\text{json}_{\text{total number of rows}}}.
\label{eq:error-rate}
\end{equation}

Lastly, to validate the proposed framework, we perform statistical hypothesis testing using McNemar's test. Statistical significance is evaluated by computing the $p$-value based on the cumulative distribution function (CDF) of the chi-squared distribution with one degree of freedom:

\begin{equation}
p = 1 - \text{CDF}\left(\chi^2 = \frac{(b - c)^2}{b + c}, \ \text{df} = 1\right)
\label{eq:pvalue}
\end{equation}

\begin{itemize}
    \item $b$ represents the number of instances correctly classified by the baseline model but misclassified after incorporating the psychology-informed module.
    \item $c$ represents the number of instances misclassified by the baseline model but correctly classified after incorporating the module.
\end{itemize}

The resulting $p$-value is compared against a conventional significance threshold $\alpha = 0.05$. If $p < 0.05$, we reject the null hypothesis, indicating that the psychology-informed module has a statistically significant impact on model performance.

\section{Discussion}

\begin{figure*}
    \centering
    \subfigure[Essays Dataset (Contents)]{
        \includegraphics[width=0.22\textwidth]{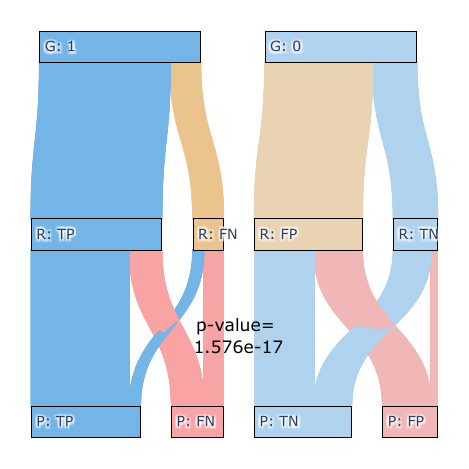}
    }
    \subfigure[Kaggle Dataset (Contents)]{
        \includegraphics[width=0.22\textwidth]{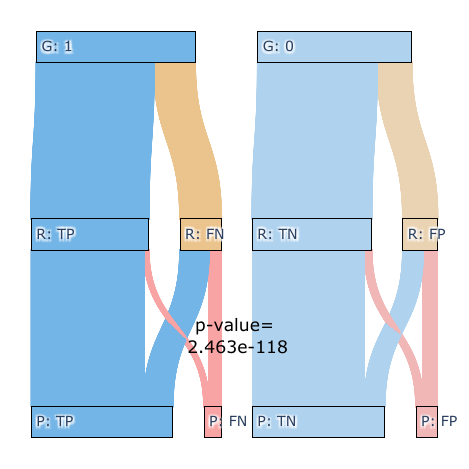}
    }
    \subfigure[Essays Dataset (Embeddings)]{
        \includegraphics[width=0.22\textwidth]{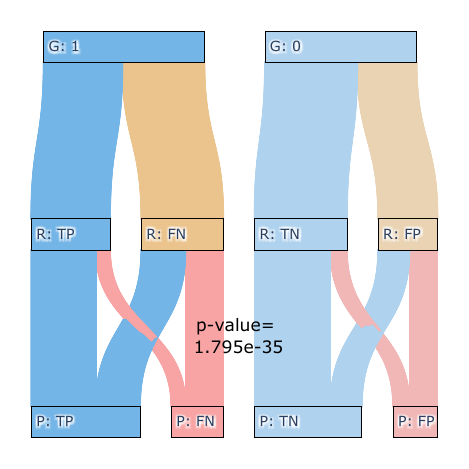}
    }
    \subfigure[Kaggle Dataset (Embeddings)]{
        \includegraphics[width=0.22\textwidth]{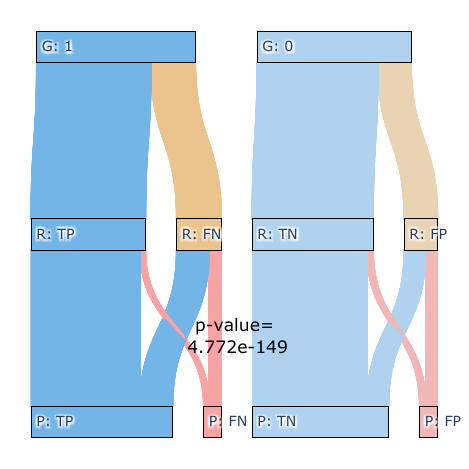}
    }
    \caption{The Sankey diagram illustrates the transition from the regular method (R) to the proposed method (P), based on the original ground truth (G), using the PICEPR approach across both the Contents Embeddings pipelines in the Essays and Kaggle datasets. We apply the McNemar p-value test to evaluate the statistical significance of the changes. Since personality has multiple dimensions, we flatten and concatenate them to reduce the complexity of the visualization. The diagram highlights that many false positives (FP) and false negatives (FN) are corrected to true positives (TP) and true negatives (TN), with fewer instances of the reverse, showcasing the effectiveness of the proposed approach.}
    \label{fig:hypo}
\end{figure*}

We conducted several experiments on the two proposed pipelines. First, Table \ref{tab:error} summarizes the error rates observed for each model during testing. Later, Table \ref{tab:llm-class-essays} and Table \ref{table:llm-class-kaggle} tabulate the results of the designed experiment for the \textbf{Contents} Pipeline. Table~\ref{tab:llm-rep-essays} and Table~\ref{tab:llm-rep-kaggle} tabulate the results of the \textbf{Embeddings} Pipeline. To further justify the effectiveness of the proposed PICEPR approach, we visualized the flow and conducted hypothesis testing after adopting the PICEPR compared to the regular method, as shown in Fig. \ref{fig:hypo}. Since the $\text{p-value} < \alpha$ in both the PICEPR Contents and Embeddings pipelines, this indicates that the approaches yields statistically significant results.

\begin{table*}[t]
\centering
\caption{PICEPR's Contents Pipeline evaluates the error rates of various LLM base models across Essays and Kaggle datasets.}
\begin{tabular}{|c|l|r|r|r|r|r|r|r|r|}
\hline
\textbf{Dataset} & \textbf{Model} & \textbf{$\mathcal{S}$} & \textbf{$\mathcal{P}$} & \textbf{$\mathcal{C}^{\text{R}}$} & \textbf{$\mathcal{C}^{\text{RT}}$} & \textbf{$\mathcal{C}^{\text{PR}}$} & \textbf{$\mathcal{C}^{\text{PR2}}$} & \textbf{$\mathcal{M}$}& \textbf{$\mathcal{V}^{\text{R}}$, $\mathcal{V}^{\text{RT}}$,$\mathcal{V}^{\text{AT}}$,$\mathcal{V}^{\text{PR}}$}\\
\hline
\multirow{5}{*}{Essays}& \textit{gpt4o} & 0 & 0 & 0 & 0 & 0 & 0 & - & 0 \\
& \textit{gemini} & 0 & 0 & 0 & 0.609 & 0 & 0 & - & 0 \\
& \textit{gpt3.5} & 0.002 & 0.006 & 0.022 & 0.032 & 0.006 & 0.028 & - & 0 \\
& llama & 0 & 0 & 0 & 0 & 0 & 0 & - & 0 \\
& mistral & 0 & 0 & 0 & 0 & 0 & 0 & - & 0 \\
\hline
\multirow{5}{*}{Kaggle} & \textit{gpt4o} & 0 & 0 & 0 & 0 & 0 & 0 & - & 0 \\
& \textit{gemini} & 0 & 0 & 0 & 0.750 & 0 & 0 & - & 0 \\
& \textit{gpt3.5} & 0.002 & 0.008 & 0.017 & 0.020 & 0.006 & 0.027 & - & 0 \\
& \textit{llama} & 0 & 0 & 0 & 0 & 0 & 0 & - & 0 \\
& \textit{mistral} & 0 & 0 & 0 & 0 & 0 & 0 & - & 0 \\
\hline
\end{tabular}
\label{tab:error}
\begin{flushleft}
\footnotesize{Note: Since this involves tabulating results only for the test set of each LLM, so \textbf{$\mathcal{M}$} is not applicable, and the \textbf{$\mathcal{S}$} only for PICEPR Contents Pipeline.
}
\end{flushleft}
\end{table*}

\begin{table*}[t]
\tiny
\centering
\caption{PICEPR's Contents Pipeline: Performance Metrics Comparison Across Models and Experiments on the Essays Dataset}
\begin{tabular}{|c|l|c|c|c|c|c|c|c|c|c|c|c|c|c|c|c|}
\hline
\multicolumn{2}{|c|}{\textbf{Experiment} } & \multicolumn{3}{|c|}{\textbf{O - Openness}} & \multicolumn{3}{|c|}{\textbf{C - Conscientiousness}} & \multicolumn{3}{|c|}{\textbf{E - Extraversion}} & \multicolumn{3}{|c|}{\textbf{A - Agreeableness}} & \multicolumn{3}{|c|}{\textbf{N - Neuroticism}} \\
\hline
Type &    Variation &      BA &      F1 &      RA &      BA &      F1 &      RA &      BA &      F1 &      RA &      BA &      F1 &      RA &      BA &      F1 &      RA \\
\hline
\multirow{5}{*}{$\mathcal{C}^{\text{R}}$} & $\{\mathcal{C},\mathcal{S},\mathcal{P}\}_{\text{gpt4o}}$ & 0.5137 & 0.6786 & 0.5283 & 0.5983 & 0.6278 & 0.5992 & 0.5521 & 0.5941 & 0.5547 & 0.5538 & 0.6837 & 0.5749 & 0.5344 & 0.6667 & 0.5344 \\
& $\{\mathcal{C},\mathcal{S},\mathcal{P}\}_{\text{gemini}}$ & 0.5319 & 0.6763 & 0.5445 & 0.5784 & 0.5217 & 0.5769 & 0.5501 & 0.5915 & 0.5526 & 0.5596 & 0.6431 & 0.5709 & 0.5283 & 0.6628 & 0.5283 \\
& $\{\mathcal{C},\mathcal{S},\mathcal{P}\}_{\text{gpt3.5}}$ & 0.5251 & 0.6799 & 0.5389 & 0.5775 & 0.5945 & 0.5779 & 0.5364 & 0.5388 & 0.5359 & 0.5359 & 0.6697 & 0.5567 & 0.5470 & 0.6429 & 0.5464 \\
& $\{\mathcal{C},\mathcal{S},\mathcal{P}\}_{\text{llama}}$ & 0.4868 & 0.6486 & 0.5000 & 0.5154 & 0.5499 & 0.5162 & 0.5686 & 0.5799 & 0.5688 & 0.5300 & 0.6636 & 0.5506 & 0.5182 & 0.6327 & 0.5182 \\
& $\{\mathcal{C},\mathcal{S},\mathcal{P}\}_{\text{mistral}}$ & 0.4888 & 0.6506 & 0.5020 & 0.5175 & 0.5475 & 0.5182 & 0.5113 & 0.5000 & 0.5101 & 0.5338 & 0.6677 & 0.5547 & 0.5061 & 0.6223 & 0.5061 \\
\hline
\multirow{5}{*}{$\mathcal{C}^{\text{RT}}$} & $\{\mathcal{C},\mathcal{S},\mathcal{P}\}_{\text{gpt4o}}$ & 0.5765 & 0.5894 & 0.5769 & 0.5660 & 0.5962 & 0.5668 & 0.5305 & 0.5650 & 0.5324 & 0.5518 & 0.5742 & 0.5526 & 0.5364 & 0.5621 & 0.5364 \\
& $\{\mathcal{C},\mathcal{S},\mathcal{P}\}_{\text{gemini}}$ & 0.5040 & 0.5479 & 0.5075 & 0.5462 & 0.5414 & 0.5448 & 0.3767 & 0.3824 & 0.3731 & 0.4683 & 0.4895 & 0.4552 & 0.4851 & 0.5106 & 0.4851\\
&$\{\mathcal{C},\mathcal{S},\mathcal{P}\}_{\text{gpt3.5}}$ & 0.5567 & 0.5657 & 0.5567 & 0.5384 & 0.5353 & 0.5381 & 0.5485 & 0.5576 & 0.5485 & 0.5318 & 0.5473 & 0.5313 & 0.5588 & 0.5731 & 0.5590 \\
& $\{\mathcal{C},\mathcal{S},\mathcal{P}\}_{\text{llama}}$ & 0.5160 & 0.5267 & 0.5162 & 0.5221 & 0.5317 & 0.5223 & 0.5793 & 0.5823 & 0.5789 & 0.5403 & 0.5720 & 0.5425 & 0.5263 & 0.5430 & 0.5263 \\
& $\{\mathcal{C},\mathcal{S},\mathcal{P}\}_{\text{mistral}}$ & 0.5203 & 0.5269 & 0.5202 & 0.5338 & 0.5594 & 0.5344 & 0.5241 & 0.5347 & 0.5243 & 0.5469 & 0.5752 & 0.5486 & 0.5142 & 0.5294 & 0.5142 \\
\hline
\multirow{5}{*}{$\mathcal{C}^{\text{PR}}$}  & $\{\mathcal{C},\mathcal{S},\mathcal{P}\}_{\text{gpt4o}}$ & 0.6667 & 0.6870 & 0.6680 & 0.7081 & \textbf{0.7198} & 0.7085 & 0.6535 & 0.6640 & 0.6538 & \textbf{0.6793} & \textbf{0.6876} & \textbf{0.6781} & \textbf{0.6721} & 0.6786 & \textbf{0.6721} \\
& $\{\mathcal{C},\mathcal{S},\mathcal{P}\}_{\text{gemini}}$ & \textbf{0.6792} & \textbf{0.6962} & \textbf{0.6802} & 0.6978 & 0.7140 & 0.6984 & 0.6360 & 0.6386 & 0.6356 & 0.6432 & 0.6495 & 0.6417 & 0.6417 & 0.6550 & 0.6417 \\
& $\{\mathcal{C},\mathcal{S},\mathcal{P}\}_{\text{gpt3.5}}$ & 0.6748 & 0.6960 & 0.6762 & 0.6940 & 0.7082 & 0.6945 & 0.6510 & 0.6667 & 0.6517 & 0.6665 & 0.6733 & 0.6653 & 0.6668 & \textbf{0.6798} & 0.6667 \\
& $\{\mathcal{C},\mathcal{S},\mathcal{P}\}_{\text{llama}}$ & 0.6567 & 0.6769 & 0.6579 & 0.6372 & 0.6538 & 0.6377 & 0.6189 & 0.6328 & 0.6194 & 0.6077 & 0.6211 & 0.6073 & 0.6154 & 0.6346 & 0.6154 \\
& $\{\mathcal{C},\mathcal{S},\mathcal{P}\}_{\text{mistral}}$ & 0.6303 & 0.6527 & 0.6316 & 0.5948 & 0.6094 & 0.5951 & 0.6070 & 0.6181 & 0.6073 & 0.5936 & 0.6067 & 0.5931 & 0.6174 & 0.6358 & 0.6174 \\
\hline
\multirow{5}{*}{$\mathcal{C}^{\text{PR2}}$} & $\{\mathcal{C},\mathcal{S},\mathcal{P}\}_{\text{gpt4o}}$ & 0.6352 & 0.6691 & 0.6377 & \textbf{0.7124} & 0.7183 & \textbf{0.7126} & \textbf{0.6654} & \textbf{0.6784} & \textbf{0.6660} & 0.6294 & 0.6335 & 0.6275 & 0.6518 & 0.6601 & 0.6518 \\
& $\{\mathcal{C},\mathcal{S},\mathcal{P}\}_{\text{gemini}}$ & 0.6372 & 0.6716 & 0.6397 & 0.7020 & 0.7146 & 0.7024 & 0.6611 & 0.6770 & 0.6619 & 0.5859 & 0.6220 & 0.5891 & 0.6073 & 0.6181 & 0.6073 \\
& $\{\mathcal{C},\mathcal{S},\mathcal{P}\}_{\text{gpt3.5}}$ & 0.6243 & 0.6617 & 0.6273 & 0.6994 & 0.7117 & 0.6998 & 0.6453 & 0.6600 & 0.6460 & 0.5881 & 0.6044 & 0.5880 & 0.6502 & 0.6586 & 0.6501  \\
& $\{\mathcal{C},\mathcal{S},\mathcal{P}\}_{\text{llama}}$ & 0.5803 & 0.6213 & 0.5830 & 0.6108 & 0.6293 & 0.6113 & 0.5452 & 0.5725 & 0.5466 & 0.5194 & 0.5564 & 0.5223 & 0.5911 & 0.5944 & 0.5911 \\
& $\{\mathcal{C},\mathcal{S},\mathcal{P}\}_{\text{mistral}}$ & 0.5716 & 0.6209 & 0.5749 & 0.5827 & 0.5961 & 0.5830 & 0.5516 & 0.5742 & 0.5526 & 0.5154 & 0.5526 & 0.5182 & 0.5870 & 0.5968 & 0.5870 \\
\hline
\end{tabular}
\label{tab:llm-class-essays}
\begin{flushleft}
\footnotesize{Note: The highest-performing algorithm for each model is highlighted in \textbf{bold}. Since this is a balanced dataset, RA and BA metrics are identical. The F1 score indicates the model's susceptibility to bias. The results for the \textit{gemini} model in $\mathcal{C}^{\text{RT}}$ are not referable due to the excessively high error rate. Notably, the fraction of valid JSON outputs that could be parsed is less than one-third of the dataset.}
\end{flushleft}

\end{table*}

\begin{table*}[t]
\centering
\caption{PICEPR's Contents Pipeline: Performance Metrics Comparison Across Models and Experiments on the Kaggle Dataset}
\begin{tabular}{|l|l|c|c|c|c|c|c|c|c|c|c|c|c|}
\hline
\multicolumn{2}{|l|}{\textbf{Experiment} } & \multicolumn{3}{|c|}{\textbf{O - Sensing/Intuition (S/N)}} & \multicolumn{3}{|c|}{\textbf{C - Judging/Perceiving (J/P)}} & \multicolumn{3}{|c|}{\textbf{E - Extra/Introversion (E/I)}} & \multicolumn{3}{|c|}{\textbf{A - Sensing/Intuition (S/N)}}  \\
\hline
Type &    Variation &      BA &      F1 &      RA &      BA &      F1 &      RA &      BA &      F1 &      RA &      BA &      F1 &      RA  \\
\hline
\multirow{5}{*}{$\mathcal{C}^{\text{R}}$} & $\{\mathcal{C},\mathcal{S},\mathcal{P}\}_{\text{gpt4o}}$ & 0.7280 & 0.8852 & 0.8115 & 0.6573 & 0.5762 & 0.6795 & 0.6990 & 0.5269 & 0.7464 & 0.7950 & 0.8015 & 0.7931 \\
& $\{\mathcal{C},\mathcal{S},\mathcal{P}\}_{\text{gemini}}$ & 0.7702 & 0.9047 & 0.8421 & 0.6721 & 0.5990 & 0.6905 & 0.7252 & 0.5498 & 0.7262 & 0.8068 & 0.8028 & 0.8017 \\
& $\{\mathcal{C},\mathcal{S},\mathcal{P}\}_{\text{gpt3.5}}$ & 0.7410 & 0.8577 & 0.7744 & 0.6488 & 0.5740 & 0.6653 & 0.6646 & 0.4837 & 0.7709 & 0.7867 & 0.7863 & 0.7826 \\
& $\{\mathcal{C},\mathcal{S},\mathcal{P}\}_{\text{llama}}$ & 0.5902 & 0.7703 & 0.6524 & 0.6339 & 0.5555 & 0.6513 & 0.6311 & 0.4360 & 0.6998 & 0.7048 & 0.6977 & 0.6996 \\
& $\{\mathcal{C},\mathcal{S},\mathcal{P}\}_{\text{mistal}}$ & 0.5827 & 0.7650 & 0.6455 & 0.5971 & 0.5267 & 0.6061 & 0.6191 & 0.4198 & 0.6948 & 0.7037 & 0.6979 & 0.6986 \\
\hline
\multirow{5}{*}{$\mathcal{C}^{\text{RT}}$} & $\{\mathcal{C},\mathcal{S},\mathcal{P}\}_{\text{gpt4o}}$ & 0.7097 & 0.8893 & 0.8161 & 0.6512 & 0.5756 & 0.6686 & 0.6497 & 0.4614 & 0.7389 & 0.7798 & 0.7892 & 0.7787 \\
& $\{\mathcal{C},\mathcal{S},\mathcal{P}\}_{\text{gemini}}$ & 0.4595 & 0.6897 & 0.5537 & 0.4139 & 0.5560 & 0.4484 & 0.5262 & 0.2917 & 0.6408 & 0.5953 & 0.5898 & 0.5953 \\
& $\{\mathcal{C},\mathcal{S},\mathcal{P}\}_{\text{gpt3.5}}$ & 0.6613 & 0.7796 & 0.6707 & 0.6375 & 0.5607 & 0.6532 & 0.6663 & 0.5013 & 0.7541 & 0.7851 & 0.7851 & 0.7808 \\
& $\{\mathcal{C},\mathcal{S},\mathcal{P}\}_{\text{llama}}$ & 0.5620 & 0.7570 & 0.6340 & 0.6289 & 0.5508 & 0.6455 & 0.6166 & 0.4160 & 0.6974 & 0.6928 & 0.6856 & 0.6876 \\
& $\{\mathcal{C},\mathcal{S},\mathcal{P}\}_{\text{mistral}}$ & 0.5639 & 0.7624 & 0.6403 & 0.5838 & 0.5147 & 0.5914 & 0.6137 & 0.4134 & 0.6876 & 0.6915 & 0.6852 & 0.6865 \\
\hline
\multirow{5}{*}{$\mathcal{C}^{\text{PR}}$}& $\{\mathcal{C},\mathcal{S},\mathcal{P}\}_{\text{gpt4o}}$ & 0.8858 & 0.9569 & 0.9268 & 0.8158 & 0.7779 & 0.8190 & 0.8579 & \textbf{0.7596} & \textbf{0.8807} & \textbf{0.9039} & \textbf{0.9091} & \textbf{0.9032} \\
& $\{\mathcal{C},\mathcal{S},\mathcal{P}\}_{\text{gemini}}$ & 0.8756 & 0.9498 & 0.9153 & 0.8069 & 0.7674 & 0.8110 & 0.8191 & 0.7066 & 0.8559 & 0.8979 & 0.9015 & 0.8963 \\
& $\{\mathcal{C},\mathcal{S},\mathcal{P}\}_{\text{gpt3.5}}$ & 0.8728 & 0.9539 & 0.9215 & 0.8177 & 0.7799 & 0.8190 & 0.8589 & 0.7565 & 0.8799 & 0.9034 & 0.9083 & 0.9021 \\
& $\{\mathcal{C},\mathcal{S},\mathcal{P}\}_{\text{llama}}$ & 0.7461 & 0.8852 & 0.8127 & 0.7455 & 0.6989 & 0.7452 & 0.8012 & 0.6712 & 0.8311 & 0.8596 & 0.8654 & 0.8582 \\
& $\{\mathcal{C},\mathcal{S},\mathcal{P}\}_{\text{mistral}}$ & 0.7411 & 0.8705 & 0.7919 & 0.7654 & 0.7208 & 0.7660 & 0.7793 & 0.6386 & 0.8121 & 0.8593 & 0.8645 & 0.8576 \\
\hline
\multirow{5}{*}{$\mathcal{C}^{\text{PR2}}$} & $\{\mathcal{C},\mathcal{S},\mathcal{P}\}_{\text{gpt4o}}$ & \textbf{0.8897} & \textbf{0.9572} & \textbf{0.9274} & \textbf{0.8223} & \textbf{0.7856} & \textbf{0.8254} & \textbf{0.8661} & 0.7444 & 0.8611 & 0.8925 & 0.8980 & 0.8916 \\
& $\{\mathcal{C},\mathcal{S},\mathcal{P}\}_{\text{gemini}}$ & 0.8785 & 0.9490 & 0.9141 & 0.8132 & 0.7747 & 0.8173 & 0.8482 & 0.7211 & 0.8484 & 0.8886 & 0.8929 & 0.8870 \\
& $\{\mathcal{C},\mathcal{S},\mathcal{P}\}_{\text{gpt3.5}}$ & 0.8834 & 0.9535 & 0.9206 & 0.8094 & 0.7690 & 0.8145 & 0.8606 & 0.7261 & 0.8524 & 0.9019 & 0.9067 & 0.9007 \\
& $\{\mathcal{C},\mathcal{S},\mathcal{P}\}_{\text{llama}}$ & 0.7609 & 0.8724 & 0.7960 & 0.7618 & 0.7169 & 0.7620 & 0.8030 & 0.6638 & 0.8190 & 0.8556 & 0.8601 & 0.8536 \\
& $\{\mathcal{C},\mathcal{S},\mathcal{P}\}_{\text{mistral}}$ & 0.7808 & 0.8722 & 0.7971 & 0.7700 & 0.7256 & 0.7712 & 0.8040 & 0.6667 & 0.8219 & 0.8576 & 0.8628 & 0.8559 \\
\hline
\end{tabular}
\label{table:llm-class-kaggle}
\begin{flushleft}
\footnotesize{ Note: The highest-performing algorithm for each model is highlighted in \textbf{bold}. As this is an imbalanced dataset, BA metrics are used to reflect overall performance, as they account for the minority class. The F1 score may not be as effective since it does not consider true negatives (in some dimensions, TN is larger than TP). The results for the gemini model in $\mathcal{C}^{\text{RT}}$ are not referable due to the excessively high error rate. Notably, the fraction of valid JSON outputs that could be parsed is less than one-third of the dataset.}
\end{flushleft}
\end{table*}

\begin{table*}[t]
\tiny
\centering
\caption{PICEPR's Embeddings Pipeline: Performance Metrics Comparison Across Models and Experiments on the Essays Dataset}
\begin{tabular}{|c|l|c|c|c|c|c|c|c|c|c|c|c|c|c|c|c|}
\hline
\multicolumn{2}{|c|}{\textbf{Experiment} } & \multicolumn{3}{|c|}{\textbf{O - Openness}} & \multicolumn{3}{|c|}{\textbf{C - Conscientiousness}} & \multicolumn{3}{|c|}{\textbf{E - Extraversion}} & \multicolumn{3}{|c|}{\textbf{A - Agreeableness}} & \multicolumn{3}{|c|}{\textbf{N - Neuroticism}} \\
\hline
Type &    Variation &      BA &      F1 &      RA &      BA &      F1 &      RA &      BA &      F1 &      RA &      BA &      F1 &      RA &BA &      F1 &      RA  \\
\hline
\multirow{6}{*}{\makecell{$\mathcal{V}^{\text{R}}$}} & $\mathcal{V}_{\text{gptada}}$ & 0.6071 & 0.6166 & 0.6073 & 0.5825 & 0.6008 & 0.5830 & 0.5815 & 0.6320 & 0.5850 & 0.5592 & 0.6645 & 0.5749 & 0.5729 & 0.5631 & 0.5729 \\
& $\mathcal{V}_{\text{gemb}}$ & 0.5785 & 0.5922 & 0.5789 & 0.5600 & 0.5898 & 0.5607 & 0.5446 & 0.6039 & 0.5486 & 0.5583 & 0.6272 & 0.5668 & 0.5405 & 0.5592 & 0.5405 \\
& $\mathcal{V}_{\text{mistend}}$ & 0.5882 & 0.6074 & 0.5891 & 0.5967 & 0.6151 & 0.5972 & 0.5572 & 0.6104 & 0.5607 & 0.5694 & 0.6314 & 0.5769 & 0.5830 & 0.5813 & 0.5830 \\
& $\mathcal{V}_{\text{minilm}}$ & 0.5930 & 0.6269 & 0.5951 & 0.5438 & 0.3747 & 0.5405 & 0.5465 & 0.4887 & 0.5425 & 0.5262 & 0.4136 & 0.5121 & 0.5486 & 0.6350 & 0.5486 \\
& $\mathcal{V}_{\text{mpnet}}$ & 0.5824 & 0.5977 & 0.5830 & 0.5784 & 0.6000 & 0.5789 & 0.5615 & 0.6112 & 0.5648 & 0.5385 & 0.6188 & 0.5486 & 0.5466 & 0.5429 & 0.5466 \\
& $\mathcal{V}_{\text{sbert}}$ & 0.5433 & 0.5388 & 0.5425 & 0.5399 & 0.5643 & 0.5405 & 0.5063 & 0.5679 & 0.5101 & 0.5333 & 0.6090 & 0.5425 & 0.5405 & 0.5377 & 0.5405 \\
\hline
\multirow{6}{*}{\makecell{$\mathcal{V}^{\text{RT}}$}} & $\mathcal{V}_{\text{minilm}}$ & 0.6069 & 0.6196 & 0.6073 & 0.5469 & 0.5372 & 0.5466 & 0.5669 & 0.6004 & 0.5688 & 0.5688 & 0.6556 & 0.5810 & 0.5789 & 0.5667 & 0.5789 \\
& $\mathcal{V}_{\text{mpnet}}$ & 0.6047 & 0.6199 & 0.6053 & 0.5704 & 0.5875 & 0.5709 & 0.5611 & 0.5909 & 0.5628 & 0.5548 & 0.6211 & 0.5628 & 0.5729 & 0.5950 & 0.5729 \\
& $\mathcal{V}_{\text{sbert}}$ & 0.5394 & 0.5328 & 0.5385 & 0.5463 & 0.5573 & 0.5466 & 0.5118 & 0.5247 & 0.5121 & 0.5398 & 0.5752 & 0.5425 & 0.5466 & 0.5466 & 0.5466 \\
& $\mathcal{V}_{\text{minilm}}$$\mathcal{P}_{\text{gpt4o}}$ & 0.5989 & 0.6102 & 0.5992 & 0.5504 & 0.5595 & 0.5506 & 0.5595 & 0.5835 & 0.5607 & 0.5837 & 0.6644 & 0.5951 & 0.5830 & 0.6241 & 0.5830 \\
& $\mathcal{V}_{\text{mpnet}}$$\mathcal{P}_{\text{gpt4o}}$ & 0.5928 & 0.6051 & 0.5931 & 0.5664 & 0.5837 & 0.5668 & 0.5568 & 0.5902 & 0.5587 & 0.5480 & 0.6191 & 0.5567 & 0.5607 & 0.5786 & 0.5607 \\
& $\mathcal{V}_{\text{sbert}}$$\mathcal{P}_{\text{gpt4o}}$ & 0.5754 & 0.5766 & 0.5749 & 0.5705 & 0.5843 & 0.5709 & 0.5427 & 0.5480 & 0.5425 & 0.5495 & 0.5869 & 0.5526 & 0.5506 & 0.5451 & 0.5506 \\
\hline
\multirow{6}{*}{\makecell{$\mathcal{V}^{\text{AT}}$}} & $\mathcal{V}_{\text{minilm}}$$\mathcal{M}'_{\text{gpt4o}}$ & 0.6398 & 0.6454 & 0.6397 & 0.6076 & 0.6008 & 0.6073 & 0.6054 & 0.6108 & 0.6053 & 0.6105 & 0.6545 & 0.6154 & 0.6134 & 0.6110 & 0.6134 \\
& $\mathcal{V}_{\text{mpnet}}$$\mathcal{M}'_{\text{gpt4o}}$ & 0.6274 & 0.6364 & 0.6275 & 0.5974 & 0.5930 & 0.5972 & 0.5996 & 0.6024 & 0.5992 & 0.6368 & 0.6788 & 0.6417 & 0.6093 & 0.6238 & 0.6093 \\
& $\mathcal{V}_{\text{sbert}}$$\mathcal{M}'_{\text{gpt4o}}$ & 0.5661 & 0.5549 & 0.5648 & 0.6035 & 0.5967 & 0.6032 & 0.5431 & 0.5425 & 0.5425 & 0.5696 & 0.5939 & 0.5709 & 0.5749 & 0.5817 & 0.5749 \\
& $\mathcal{V}_{\text{minilm}}$$\mathcal{M}'_{\text{mistral}}$ & 0.6015 & 0.6052 & 0.6012 & 0.5445 & 0.5473 & 0.5445 & 0.5313 & 0.5549 & 0.5324 & 0.5347 & 0.6021 & 0.5425 & 0.5445 & 0.5856 & 0.5445 \\
& $\mathcal{V}_{\text{mpnet}}$$\mathcal{M}'_{\text{mistral}}$ & 0.5894 & 0.5932 & 0.5891 & 0.5462 & 0.5608 & 0.5466 & 0.5411 & 0.5687 & 0.5425 & 0.5582 & 0.6384 & 0.5688 & 0.5466 & 0.5556 & 0.5466 \\
& $\mathcal{V}_{\text{sbert}}$$\mathcal{M}'_{\text{mistral}}$ & 0.5582 & 0.5428 & 0.5567 & 0.5345 & 0.5325 & 0.5344 & 0.5184 & 0.5537 & 0.5202 & 0.5232 & 0.5618 & 0.5263 & 0.5385 & 0.5494 & 0.5385 \\
\hline

\multirow{6}{*}{\makecell{$\mathcal{V}^{\text{PR}}$}} & $\mathcal{V}_{\text{minilm}}$$\{\mathcal{S},\mathcal{M}\}_{\text{gpt4o}}$ & \textbf{0.7443} & \textbf{0.7381} &\textbf{0.7429} & \textbf{0.7001} & \textbf{0.7098} & \textbf{0.7004} & \textbf{0.6894} & \textbf{0.6844} & \textbf{0.6883} & 0.6504 & 0.6730 & 0.6518 & \textbf{0.6741} & \textbf{0.6596} & \textbf{0.6741} \\
 & $\mathcal{V}_{\text{mpnet}}$$\{\mathcal{S},\mathcal{M}\}_{\text{gpt4o}}$ & 0.7278 & 0.7031 & 0.7247 & 0.6976 & 0.7183 & 0.6984 & 0.6679 & 0.6759 & 0.6680 & \textbf{0.6530} & \textbf{0.7032} & \textbf{0.6599} & 0.6538 & 0.6415 & 0.6538 \\
 & $\mathcal{V}_{\text{sbert}}$$\{\mathcal{S},\mathcal{M}\}_{\text{gpt4o}}$ & 0.6636 & 0.6514 & 0.6619 & 0.6012 & 0.6036 & 0.6012 & 0.6197 & 0.6478 & 0.6215 & 0.6201 & 0.6314 & 0.6194 & 0.6417 & 0.6467 & 0.6417 \\
 & $\mathcal{V}_{\text{minilm}}$$\{\mathcal{S},\mathcal{M}\}_{\text{mistral}}$ & 0.7083 & 0.6960 & 0.7065 & 0.6604 & 0.6485 & 0.6599 & 0.6492 & 0.6628 & 0.6498 & 0.6609 & 0.6586 & 0.6579 & 0.6680 & 0.6733 & 0.6680 \\
 & $\mathcal{V}_{\text{mpnet}}$$\{\mathcal{S},\mathcal{M}\}_{\text{mistral}}$ & 0.6826 & 0.6624 & 0.6802 & 0.6279 & 0.6183 & 0.6275 & 0.6271 & 0.6392 & 0.6275 & 0.6183 & 0.6154 & 0.6154 & 0.6579 & 0.6613 & 0.6579 \\
 & $\mathcal{V}_{\text{sbert}}$$\{\mathcal{S},\mathcal{M}\}_{\text{mistral}}$ & 0.6682 & 0.6497 & 0.6660 & 0.5873 & 0.5802 & 0.5870 & 0.6169 & 0.6301 & 0.6174 & 0.5993 & 0.6012 & 0.5972 & 0.6336 & 0.6472 & 0.6336 \\
\hline
\end{tabular}
\label{tab:llm-rep-essays}
\begin{flushleft}
\footnotesize{Note: The highest-performing algorithm for each model is highlighted in \textbf{bold}. Since this is a balanced dataset, RA and BA metrics are identical.}
\end{flushleft}
\end{table*}

\begin{table*}[t]
\scriptsize
\centering
\caption{PICEPR's Embeddings Pipeline: Performance Metrics Comparison Across Models and Experiments on the Kaggle Dataset}
\begin{tabular}{|c|l|c|c|c|c|c|c|c|c|c|c|c|c|}
\hline
\multicolumn{2}{|c|}{\textbf{Experiment} } & \multicolumn{3}{|c|}{\textbf{O - Sensing/Intuition (S/N)}} & \multicolumn{3}{|c|}{\textbf{C - Judging/Perceiving (J/P)}} & \multicolumn{3}{|c|}{\textbf{E - Extra/Introversion (E/I)}} & \multicolumn{3}{|c|}{\textbf{A - Sensing/Intuition (S/N)}}  \\
\hline
Type &    Variation &      BA &      F1 &      RA &      BA &      F1 &      RA &      BA &      F1 &      RA &      BA &      F1 &      RA  \\
\hline
\multirow{6}{*}{\makecell{$\mathcal{V}^{\text{R}}$}} & $\mathcal{V}_{\text{gptada}}$ & 0.7798 & 0.9011 & 0.8375 & 0.7042 & 0.6350 & 0.7251 & 0.7275 & 0.5705 & 0.7821 & 0.8262 & 0.8444 & 0.8288 \\
& $\mathcal{V}_{\text{gemb}}$& 0.7788 & 0.8980 & 0.8329 & 0.7306 & 0.6702 & 0.7481 & 0.6305 & 0.4418 & 0.6156 & 0.8365 & 0.8526 & 0.8386 \\
& $\mathcal{V}_{\text{mistend}}$ & 0.7229 & 0.8899 & 0.8179 & 0.7327 & 0.6713 & 0.7522 & 0.7113 & 0.5490 & 0.7746 & 0.8522 & 0.8653 & 0.8536 \\
& $\mathcal{V}_{\text{minilm}}$ & 0.7524 & 0.8946 & 0.8265 & 0.6315 & 0.5242 & 0.6663 & 0.6297 & 0.4310 & 0.7337 & 0.7774 & 0.8004 & 0.7804 \\
& $\mathcal{V}_{\text{mpnet}}$ &0.7612 & 0.9026 & 0.8386 & 0.6364 & 0.5330 & 0.6697 & 0.6316 & 0.4332 & 0.7406 & 0.7869 & 0.8102 & 0.7902 \\
& $\mathcal{V}_{\text{sbert}}$ &0.7574 & 0.9004 & 0.8352 & 0.6342 & 0.5305 & 0.6674 & 0.6131 & 0.4100 & 0.7014 & 0.7845 & 0.8065 & 0.7873 \\
\hline
\multirow{6}{*}{\makecell{$\mathcal{V}^{\text{RT}}$}} & $\mathcal{V}_{\text{minilm}}$ & 0.7791 & 0.9086 & 0.8484 & 0.6395 & 0.5460 & 0.6674 & 0.6520 & 0.4647 & 0.7331 & 0.7916 & 0.8108 & 0.7937 \\
& $\mathcal{V}_{\text{mpnet}}$ & 0.7545 & 0.9033 & 0.8392 & 0.6420 & 0.5449 & 0.6726 & 0.6431 & 0.4521 & 0.7233 & 0.7958 & 0.8176 & 0.7988 \\
& $\mathcal{V}_{\text{sbert}}$ & 0.7540 & 0.8986 & 0.8323 & 0.6261 & 0.5233 & 0.6576 & 0.6336 & 0.4389 & 0.7141 & 0.7908 & 0.8124 & 0.7937 \\
& $\mathcal{V}_{\text{minilm}}$$\mathcal{P}_{\text{gpt4o}}$ & 0.7251 & 0.8966 & 0.8277 & 0.6400 & 0.5494 & 0.6663 & 0.6509 & 0.4631 & 0.7314 & 0.7943 & 0.8137 & 0.7965 \\
& $\mathcal{V}_{\text{mpnet}}$$\mathcal{P}_{\text{gpt4o}}$ & 0.7220 & 0.8951 & 0.8254 & 0.6439 & 0.5466 & 0.6749 & 0.6436 & 0.4528 & 0.7228 & 0.7879 & 0.8098 & 0.7908 \\
& $\mathcal{V}_{\text{sbert}}$$\mathcal{P}_{\text{gpt4o}}$ & 0.7441 & 0.8934 & 0.8242 & 0.6348 & 0.5353 & 0.6657 & 0.6402 & 0.4480 & 0.7216 & 0.7932 & 0.8143 & 0.7960 \\
\hline
\multirow{6}{*}{\makecell{$\mathcal{V}^{\text{AT}}$}} & $\mathcal{V}_{\text{minilm}}$$\mathcal{M}'_{\text{gpt4o}}$ & 0.8214 & 0.9177 & 0.8640 & 0.7410 & 0.6797 & 0.7631 & 0.8108 & 0.6822 & 0.8352 & 0.8490 & 0.8604 & 0.8496 \\
& $\mathcal{V}_{\text{mpnet}}$$\mathcal{M}'_{\text{gpt4o}}$ &0.8079 & 0.9147 & 0.8588 & 0.7366 & 0.6730 & 0.7602 & 0.7951 & 0.6600 & 0.8231 & 0.8491 & 0.8602 & 0.8496 \\
& $\mathcal{V}_{\text{sbert}}$$\mathcal{M}'_{\text{gpt4o}}$ &0.7927 & 0.9179 & 0.8628 & 0.7133 & 0.6456 & 0.7349 & 0.7881 & 0.6511 & 0.8190 & 0.8406 & 0.8518 & 0.8409 \\
& $\mathcal{V}_{\text{minilm}}$$\mathcal{M}'_{\text{mistral}}$ &0.7681 & 0.9043 & 0.8415 & 0.6864 & 0.6120 & 0.7084 & 0.7804 & 0.6357 & 0.8058 & 0.8024 & 0.8161 & 0.8029 \\
& $\mathcal{V}_{\text{mpnet}}$$\mathcal{M}'_{\text{mistral}}$ &0.7644 & 0.9021 & 0.8380 & 0.6822 & 0.6047 & 0.7061 & 0.7736 & 0.6258 & 0.7994 & 0.7971 & 0.8114 & 0.7977 \\
& $\mathcal{V}_{\text{sbert}}$$\mathcal{M}'_{\text{mistral}}$ &0.7609 & 0.9003 & 0.8352 & 0.6659 & 0.5867 & 0.6882 & 0.7533 & 0.5991 & 0.7856 & 0.7924 & 0.8054 & 0.7925 \\
\hline
\multirow{6}{*}{\makecell{$\mathcal{V}^{\text{PR}}$}} & $\mathcal{V}_{\text{minilm}}$$\{\mathcal{S},\mathcal{M}\}_{\text{gpt4o}}$ &\textbf{0.8731} & 0.9273 & 0.8807 & \textbf{0.8563} & \textbf{0.8254} & 0.8576 & \textbf{0.8875} & \textbf{0.7817} & \textbf{0.8847} & \textbf{0.9276} & \textbf{0.9333} & \textbf{0.9280} \\
& $\mathcal{V}_{\text{mpnet}}$$\{\mathcal{S},\mathcal{M}\}_{\text{gpt4o}}$ &0.8605 & \textbf{0.9353} & \textbf{0.8922} & 0.8541 & 0.8233 & \textbf{0.8582} & 0.8780 & 0.7579 & 0.8674 & 0.9162 & 0.9225 & 0.9164 \\
& $\mathcal{V}_{\text{sbert}}$$\{\mathcal{S},\mathcal{M}\}_{\text{gpt4o}}$ & 0.7966 & 0.9181 & 0.8634 & 0.8173 & 0.7792 & 0.8265 & 0.8389 & 0.7126 & 0.8461 & 0.8951 & 0.8999 & 0.8939 \\
& $\mathcal{V}_{\text{minilm}}$$\{\mathcal{S},\mathcal{M}\}_{\text{mistral}}$ & 0.8257 & 0.9245 & 0.8744 & 0.8310 & 0.7957 & 0.8352 & 0.8427 & 0.7158 & 0.8467 & 0.8905 & 0.8967 & 0.8899 \\
& $\mathcal{V}_{\text{mpnet}}$$\{\mathcal{S},\mathcal{M}\}_{\text{mistral}}$ &0.7876 & 0.9142 & 0.8571 & 0.8295 & 0.7940 & 0.8334 & 0.8391 & 0.7108 & 0.8438 & 0.8803 & 0.8867 & 0.8795 \\
& $\mathcal{V}_{\text{sbert}}$$\{\mathcal{S},\mathcal{M}\}_{\text{mistral}}$ &0.7876 & 0.9142 & 0.8571 & 0.8295 & 0.7940 & 0.8334 & 0.8391 & 0.7108 & 0.8438 & 0.8803 & 0.8867 & 0.8795 \\
\hline
\end{tabular}
\label{tab:llm-rep-kaggle}
\begin{flushleft}
\footnotesize{Note: The highest-performing algorithm for each model is highlighted in \textbf{bold}. As this is an imbalanced dataset, BA metrics are used to reflect overall performance as they account for the minority class. The F1 score may not be as effective for evaluation since it does not consider true negatives (particularly for the `C' and `E' dimensions, where TN is larger than TP).}
\end{flushleft}
\end{table*}

\subsection{Invalid Generated Output}
 For the $\mathcal{V}$ LLM, the error rate is 0 because it is an encoder-only model that outputs embeddings, which inherently do not produce errors. The same observation applies to the $\mathcal{S}$ and $\mathcal{P}$ LLMs, as they have relatively simple structures output. The $\mathcal{M}$ LLM is not applicable for this analysis. However, it is worth mentioning that several errors were observed in the training set due to tokenization length issues. Therefore, we cannot conclude that the structured output engine can be fully relied upon, as it may fail to process the entire script to the end of the token (eg. [EOS] for OpenAI model).

The $\mathcal{C}$ LLM is an output component for the PICEPR Contents Pipeline.  $\mathcal{C}_{\text{gpt4o}}$, $\mathcal{C}_{\text{gemini}}$, $\mathcal{C}_{\text{llama}}$, and $\mathcal{C}_{\text{mistral}}$, there is 0\% error due to the structured output engine being enabled.

However, $\mathcal{C}_{\text{gpt3.5}}$ exhibits some errors because its API does not support a structured output engine. However, a notable trend can be observed: the error rate was significantly higher for the $\mathcal{P}$ model, which requires generating personality facets, a more complex JSON structure, than for the $\mathcal{S}$ model, which mainly involves summarizing information, including some CoT outputs.

An interesting finding is that the error rate decreased for both datasets when using the PICEPR method. This improvement can be attributed to replacing the lengthy original input with a summarized version, allowing the decoder to adhere better to its system instructions during prompting. This also explains why the error rate decreased for $\mathcal{C}^{\text{PR}}_{\text{gpt3.5}}$, but increased again when longer and more complex inputs were provided in $\mathcal{C}^{\text{PR2}}_{\text{gpt3.5}}$.

On the other hand, an exception arises with $\mathcal{C}^{\text{RT}}_{\text{gemini}}$, as it is a closed source model and its API does not support structured output for fine-tuned models. This results in an error rate of 60\% for $\mathcal{C}^{\text{RT}}_{\text{gemini}}$ and up to 75\% for MBTI tasks, indicating that more than half of the dataset is missed. Consequently, the respective results are no longer statistically significant.

\subsection{Contents Pipeline}

 In general, $\mathcal{C}_{\text{gpt4o}}$, $\mathcal{C}_{\text{gemini}}$, and $\mathcal{C}_{\text{gpt3.5}}$ outperform $\mathcal{C}_{\text{llama}}$ and $\mathcal{C}_{\text{mistral}}$ in both Essays and Kaggle Dataset.

\subsubsection{Regular CoT Prompt, $\mathcal{C}^R$}
While we experimented with zero-shot and 2-shot prompting, the performance improvements were not significant. Moreover, n-shot prompting consumed substantially more tokens—approximately $n$ times as many—without yielding proportionate improvements. This results do not show significant improvement compared to regular personality recognition approaches (eg. BERT). Moreover, they are consistent with the findings reported in \cite{li2024eerpdleveragingemotionemotion} and \cite{Hu_He_Wang_Zhao_Shao_Nie_2024}.

In the Essays dataset, we observe that the Balanced Accuracy (BA) and Regular Accuracy (RA) are identical due to the dataset's balanced nature. However, the F1 score is generally higher than both RA and BA. This indicates that the model exhibits a bias in outputting certain labels more frequently. In contrast, the Kaggle dataset, where the ``Thinking/Feeling (T/F)" dimension is relatively balanced, does not display such bias in label output. This difference can be attributed to the labeling strategies employed in prompting. In the Essays dataset, we prompt the language model to output binary labels (e.g., $0$ or $1$ for Agreeableness) for each dimension. Meanwhile, in the Kaggle dataset, we ask the model to output labels according to the MBTI format (e.g., `T/F' for ``Agreeableness"). The key distinction lies in the interpretability of the labels. The binary labels in the Big-5 structure lack inherent semantic meaning and rely solely on the label ``Agreeableness," which does not explicitly convey ``positive" or ``negative". In contrast,  in the MBTI theory, labels such as ``Thinking/Feeling" or `T/F' are meaningful tokens that the model has likely encountered during pretraining, enabling it to better contextualize and predict these labels. This lead the model to hallucinate during decoding by assigning arbitrary associations to these labels.

Moving on, the LLM demonstrates a tendency toward bias, particularly when producing labels for imbalanced datasets (e.g., Kaggle). This can be attributed to the LLM over-exploring certain personality types during pretraining, as individuals with these personalities are more likely to produce written text on the internet which aligns with the way the Kaggle dataset was collected.

\subsubsection{Regular CoT Prompt (Fine-Tuned LLM), $\mathcal{C}^{\text{RT}}$}
Fine-tuning the LLM does not lead to significant improvements and, in some cases, results in a decrement in RA/BA performance. For example, in the Conscientiousness (C) dimension of the Essays dataset, $\mathcal{C}_{\text{gemini}}$ showed reduced RA/BA after fine-tuning. While fine-tuning ensures consistency in RA/BA, it also causes a reduction in the F1 score. This demonstrates that fine-tuning enables the LLM to interpret the meaning of the output ($0$ or $1$) in the context of MBTI labels. This effect is particularly noticeable in the Openness (O), Extraversion (E), and Agreeableness (A) dimensions of the Essays dataset. However, in the Kaggle dataset, where the natural labels inherently provide insight for classification (e.g., MBTI labels), a similar decrement in RA/BA is observed despite the meaningfulness of the labels. These results suggest that fine-tuning a pretrained LLM may not be worthwhile. It demands additional computational resources but offers only limited improvement, making it less cost-effective.

\subsubsection{PICEPR, $\mathcal{C}^{\text{PR}}$}
The proposed algorithm demonstrates significant improvement (over 10\%) compared to regular Chain-of-Thought (CoT) prompting. This is verified across two datasets. The F1 score and RA/BA metrics in the Essays dataset are nearly identical, indicating that the algorithm effectively resolves biased results that appeared in regular CoT methods. Moreover, a particularly interesting finding is that, even in imbalanced datasets, the LLM model tends to produce unbiased results, as evidenced by the reduced difference between RA and BA metrics.

We experimented with a fine-tuned version of the LLM and applied the PICEPR algorithm, observing a slight decrement in performance across both datasets. This decline may be attributed to the repeated feeding of the LLM's outputs back into itself for training, which likely reinforces preexisting shortcomings. Conversely, when employing a series-based prompting approach, the feedback loop facilitates iterative refinement, as each step prompts the LLM to further extract psychological or cognitive activity from the text. Empirical results for continuing conversations yield similar outcomes, although this modularised approach consumes more tokens due to the inclusion of previous input messages.

\subsubsection{PICEPR (2-shot prompt), $\mathcal{C}^{\text{PR2}}$}
When implementing 2-shot prompting, the bias performance improves further in imbalanced datasets. This is particularly evident in the Openness dimension of the Kaggle dataset, which exhibits an 80:20 label ratio in the "Sensing/Intuition (S/N)" dimension. The improvement can be attributed to the model being exposed to both possible outputs and corresponding examples. Nevertheless, the performance does not improve for balanced datasets, with some dimensions showing a slight decline (e.g., the Openness (O), Extraversion (E), and Neuroticism (N) dimensions in the Essays dataset and "Thinking/Feeling (T/F)" in the Kaggle dataset). This indicates that while providing examples can aid in certain scenarios, there is no conclusive evidence that the 2-shot approach consistently improves performance for decoder-only LLMs.

\subsection{Embeddings Pipeline}

In general, the \textit{gpt4o} LLM ($\mathcal{S}_{\text{gpt4o}}$ \& $\mathcal{M}_{\text{gpt4o}}$) exhibits higher accuracy, while the choice of encoder model ($\mathcal{V})$ has minimal impact.

\subsubsection{Regular Encoder Classification,$\mathcal{V}^{\text{R}}$}
Compared to the results from the Contents Pipeline, the performance is significantly better, with an average improvement of approximately 5\% and avoidance of random guessing in any dimension. However, the presented result still indicate that regular NLU models are not well-suited for personality classification tasks. Meanwhile, $\mathcal{V}^{\text{R}}_{\text{gptada}}$ and $\mathcal{V}^{\text{R}}_{\text{mistemd}}$ demonstrate relatively strong representation capabilities. When comparing the RA results on the Kaggle dataset for both pipelines, this pipeline also reveals that the encoder model exhibits a relative bias.

\subsubsection{Regular Encoder Fine-Tuning, $\mathcal{V}^{\text{RT}}$}
We were unable to fine-tune closed-source encoder models, as they do not support fine-tuning due to limitations in their application programming interfaces. Consequently, we only examined open-source models for encoder fine-tuning.

By using the Essays and Kaggle datasets for fine-tuning, the results indeed improved; however, the improvement is limited. This limitation arises due to the small size of the training set, which hinders the model's ability to generalize effectively. Notably, this is evident in the Kaggle dataset, where the performance improvement is significantly higher compared to the Essays dataset. Additionally, observations from the Kaggle dataset indicate that fine-tuning indeed mitigates the dataset imbalance issue more effectively compared to $\mathcal{V}^{\text{R}}$.

Moreover, we observed that the presence of LLMs within $\mathcal{P}$ only yields improvements for $\mathcal{V}^{\text{RT}}_{\text{sbert}}$ (where the encoder underperforms), whereas $\mathcal{V}^{\text{RT}}_{\text{minilm}}$ and $\mathcal{V}^{\text{RT}}_{\text{mpnet}}$ result in slight performance degradation. This suggests that when the encoder alone is sufficient to capture personality-related information, and the additional input from the decoder-based LLMs may disrupt its focus. Although the MLP is capable of adjusting weights to some extent, the LLM decoder sometimes introduce hallucinations (that are not stable across any personality facets), causing the MLP to fail in effectively reconciling the information. Hence, we omit $\mathcal{P}$ in the subsequent experiments.

\subsubsection{LLM Augmentation Fine-Tuning, $\mathcal{V}^{\text{AT}}$}

From the results, the augmented data contains meaningful content and manages to improve the encoder's performance, allowing it to focus on differentiating personality traits rather than merely capturing the semantic meaning of sentences through a contractive loss. We utilized the closed-source $\mathcal{M}'_{\text{gpt4o}}$ model and the open-source $\mathcal{M}'_{\text{mistral}}$ model, as the latter demonstrated relatively good performance in previous experiments. As expected, the BA results (especially the most imbalance `O' dimension in Kaggle dataset) demonstrate that it inherits the advantages of fine-tuning - addressing the class imbalance issue. In addition, the RA results for $\mathcal{V}^{\text{AT}}_{\text{mistral}}$ show limited improvement (with some instances still degraded), as does $\mathcal{V}^{\text{AT}}_{\text{gpt4o}}$ compared to $\mathcal{V}^{\text{RT}}$. This can be attributed to $\mathcal{V}^{\text{AT}}_{\text{mistral}}$'s tendency to generate hallucinated content due to the length of the decoder sequence. For instance, a typical decoder classification involves a system prompt (approximately 500 tokens), user input (approximately 1000 tokens), and augmentation content (approximately 500 tokens), with some CoT reasoning interspersed. The long token length induces the lost-in-the-middle phenomenon in LLMs \cite{liu2023lostmiddlelanguagemodels}.

\subsubsection{PICEPR, $\mathcal{V}^{\text{PR}}$}
We proposed PICEPR, which the framework assigns specific roles to LLMs for personality recognition tasks to resolve the limitations of $\mathcal{V}^{\text{AT}}$. This includes $\mathcal{S}$ for summarization and $\mathcal{M}$ for content generation. By adopting this role-based approach, PICEPR enables more effective chain-of-thought (CoT) summarization and content generation, thereby mitigating the challenges associated with lengthy decoding sequences.

\begin{figure*}

    \centering
    \subfigure[Essays Dataset (5 Personality Dimensions)]{
        \includegraphics[width=0.48\textwidth]{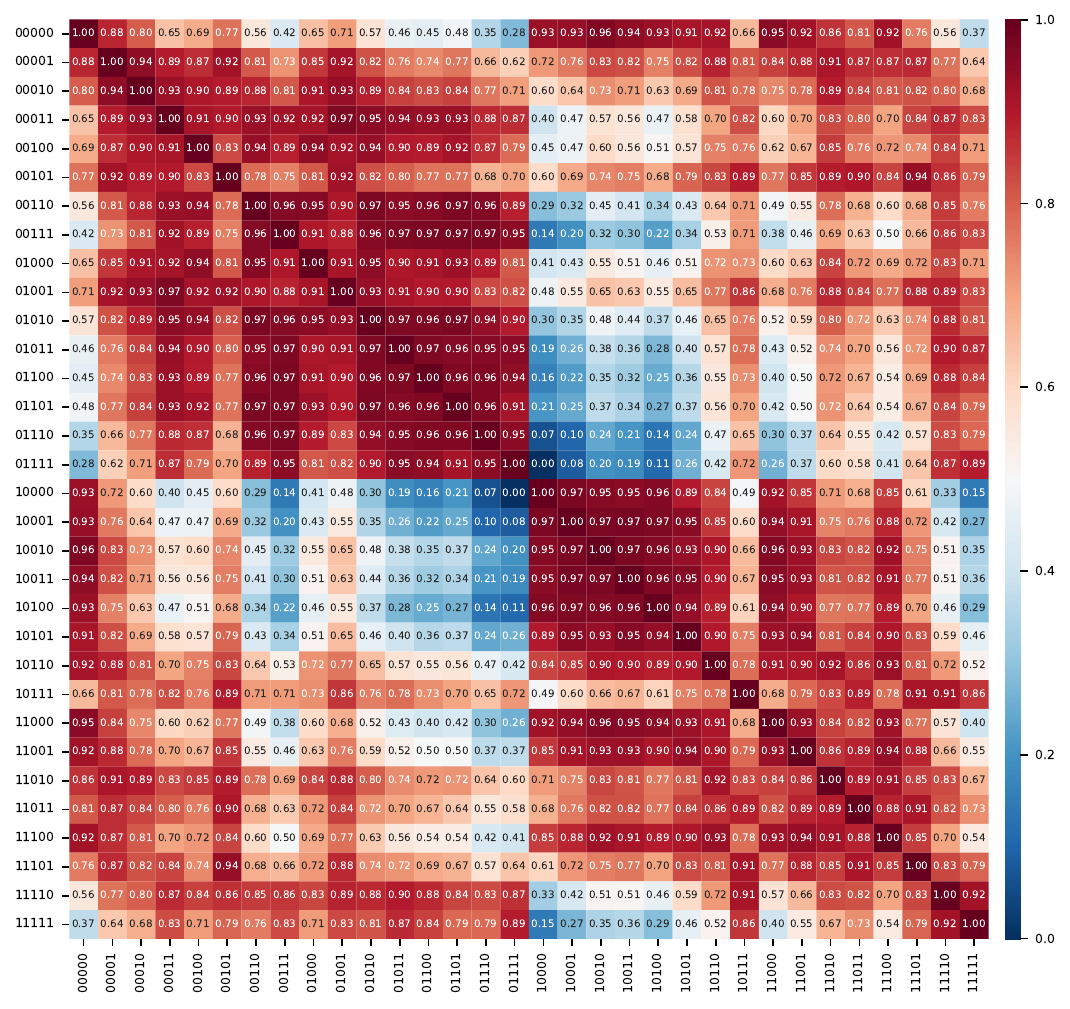}
    }
    \subfigure[Kaggle Dataset (4 Personality Dimensions)]{
        \includegraphics[width=0.48\textwidth]{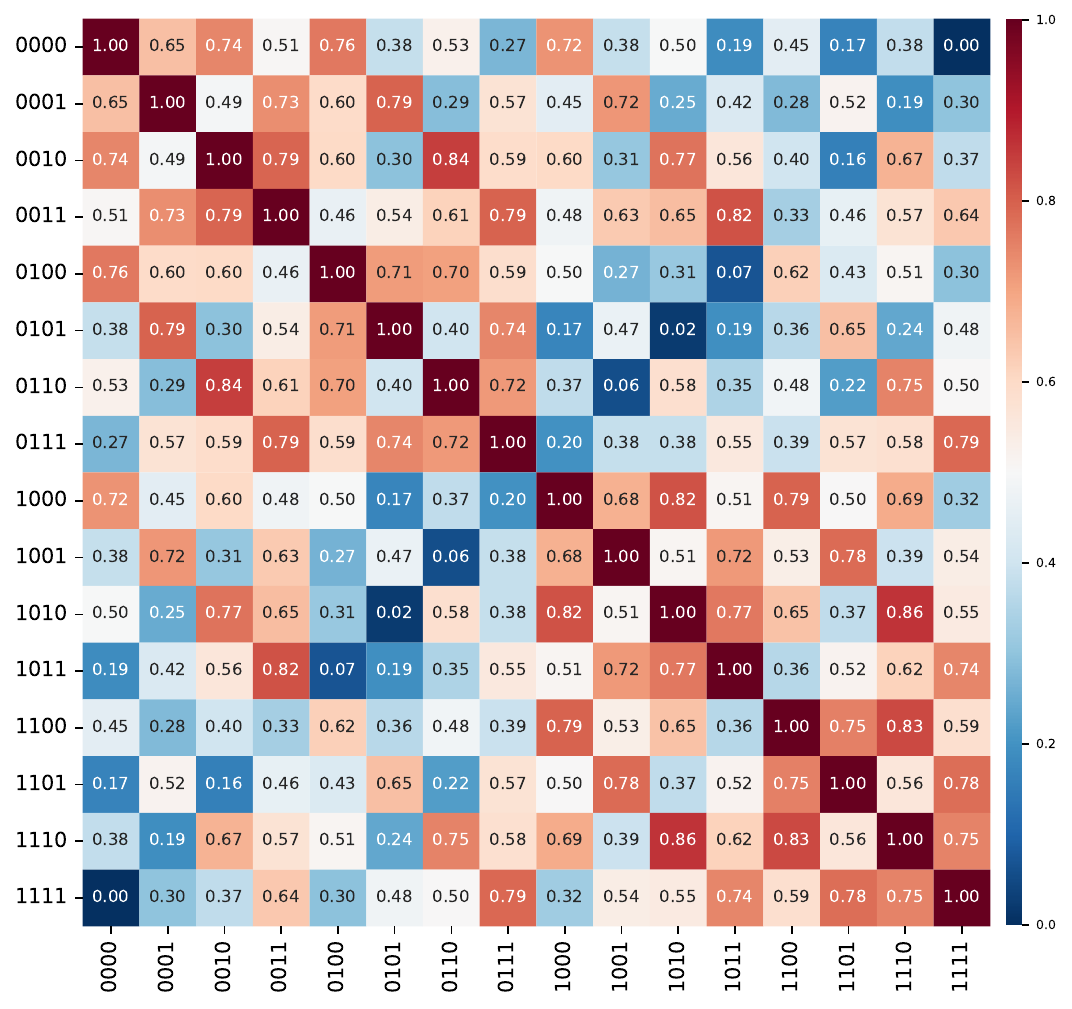}
    }
    \caption{Heatmap of normalized embedding distances by $\mathcal{V}^{\text{PR}}_{\text{minilm}}$ after the PICEPR Embeddings Pipeline (averaged) on both the Essays and Kaggle datasets. Greater differences between embeddings result in larger distances. Significant variation across personality dimensions also contributes to these distances—for example, `00000' and `00001' are close, while both are far from `11111'; similarly, `10011' and `10111' are relatively close compared to `01100'. The Kaggle dataset better preserves meaningful structure, showing consistent distance changes for single-dimension flips across any position. In contrast, the Essays dataset shows less differentiation for certain dimensions, indicating weaker separation in the embedding space.}
    
    \label{fig:similarity}
\end{figure*}

We observed that the types of decoder models, $\mathcal{S}$ and $\mathcal{M}$, played a crucial role in affecting the performance of $\mathcal{V}^{\text{PR}}_{\text{gpt4o}}$. As shown in Figure \ref{fig:similarity}, we visualized how the similarity algorithm influences the encoder model to produce embeddings that reflect personality traits. When personalities are similar (i.e., sharing more labels across personality dimensions), the resulting embeddings tend to be more alike, resulting in higher cosine similarity. Consequently, this can influence the behavior of the backbone model, which was originally designed to differentiate sentence meanings rather than personality traits. As a result, we observed an improvement in balanced accuracy (BA) in the imbalanced dimensions of the Kaggle dataset—particularly in the MBTI's `O' (Openness) dimension, which is the most imbalanced class. This demonstrates that the fine-tuning process builds a strong foundation for personality classification.

Additionally, the choice of encoder is crucial, as $\mathcal{V}^{\text{PR}}_{\text{minilm}}$ not only generally outperforms $\mathcal{V}^{\text{PR}}_{\text{sbert}}$ and $\mathcal{V}^{\text{PR}}_{\text{mpnet}}$ despite being 4 times smaller but also converges around 3.5 times faster, while achieving better performance. This underscores the importance of a well-designed and properly pre-trained model. An effective architecture outweighs the significance of the sheer number of hyperparameters for personality recognition.

\subsection{Work Comparison}
The proposed PICEPR algorithm outperforms current state-of-the-art models for both the LLM-based approach and regular deep learning methods, as tabulated in Table \ref{tab:essays-performance} and Table \ref{tab:kaggle-performance}. Compared to regular machine learning methods, such as those proposed in \cite{Amirhosseini2020, yang-etal-2021-psycholinguistic, yang2023orders, Yang_Quan_Yang_Yu_2021, Hu_He_Wang_Zhao_Shao_Nie_2024}, PICEPR achieves up to a 20\% improvement in performance. When compared to the LLM-based approach described in \cite{li2024eerpdleveragingemotionemotion}, PICEPR demonstrates an additional improvement of up to 10\%. A notable drawback exists for the PICEPR (Embeddings Pipeline) with respect to the `O' dimension in the Kaggle class RA score. However, using a regular method, the RA score for the `O' dimension could reach up to 0.9378, surpassing the performance reported in \cite{li2024eerpdleveragingemotionemotion}, which employed an LLM-based approach. Nevertheless, to have fair comparison, we aim to address this issue due to the class imbalance, which none of the previous LLM-based works have attempted to tackle. This highlights a limitation in RA metric optimization. For other classes, it is evident that even with class imbalance considered, our method still surpasses prior metrics, demonstrating the effectiveness of PICEPR.

\begin{table*}[t]
\tiny
\centering
\caption{Essays Dataset: Performance Comparison}
\begin{tabular}{|l|c|c|c|c|c|c|c|c|c|c|c|c|c|c|c|}
\hline
\multirow{2}{*}{\textbf{Works}} & \multicolumn{3}{|c|}{\textbf{O - Openness}} & \multicolumn{3}{|c|}{\textbf{C - Conscientiousness}} & \multicolumn{3}{|c|}{\textbf{E - Extraversion}} & \multicolumn{3}{|c|}{\textbf{A - Agreeableness}} & \multicolumn{3}{|c|}{\textbf{N - Neuroticism}} \\
\cline{2-16}
&      BA &      F1 &      RA &      BA &      F1 &      RA &      BA &      F1 &      RA &      BA &      F1 &      RA & BA &      F1 &      RA  \\
\hline
Word2Vec + Psycholinguistic \cite{Majumder2017} & - & - & 0.6113 & - &   & 0.5671 & - &   & 0.5809 &   &   & 0.5671 & - &   & 0.5733\\
\hline
 Psycholinguistic MLP \cite{Mehta2020} &   &   & 0.6460 &   &   & 0.5920 &   &   & 0.6000 &   &   & 0.5880 &   &   & 0.6050\\
\hline
BERT MLP \cite{Mehta2020} & - & - & 0.6040 &   & - & 0.5730 &   & - & 0.5690 &   & - & 0.5700 &   & - & 0.5980\\
\hline
CoT with Emotion \cite{li2024eerpdleveragingemotionemotion}& - & 0.6093 & 0.6102 & - & \textbf{0.6864} & 0.6800 & - & 0.6302 & 0.6201 & - & 0.6501 & 0.6498 & - & 0.5600 & 0.5600\\
\hline
\textbf{PICEPR (Contents Pipeline)} & \textbf{0.6667} & \textbf{0.6870*} & \textbf{0.6680} & \textbf{0.7081*} & \textbf{0.7198*} & \textbf{0.7085*} & \textbf{0.6535} & \textbf{0.6640*} & \textbf{0.6538} & \textbf{0.6793} & \textbf{0.6876} & \textbf{0.6781} & \textbf{0.6721} & \textbf{0.6786*} & \textbf{0.6721}\\
\hline
\textbf{PICEPR (Embeddings Pipeline)} & \textbf{0.7316*} & \textbf{0.6868} & \textbf{0.7267*} & \textbf{0.6920} & 0.6417 & \textbf{0.6903} & \textbf{0.6592*} & \textbf{0.6515} & \textbf{0.6579*} & \textbf{0.6800*} & \textbf{0.7054*} & \textbf{0.6822*} & \textbf{0.6822*} & \textbf{0.6488} & \textbf{0.6822*}\\
\hline
\end{tabular}
\label{tab:essays-performance}
\begin{flushleft}
\footnotesize{Note: The \textbf{bold} text indicates the second-highest performance, while the \textbf{bold} text with an asterisk (\textbf{*}) represents the highest performance among all other algorithms. Since this is a balanced dataset, RA and BA metrics are identical.}
\end{flushleft}
\end{table*}

\begin{table*}[t]
\tiny
\centering
\caption{Kaggle Dataset: Performance Comparison}
\begin{tabular}{|l|c|c|c|c|c|c|c|c|c|c|c|c|}
\hline
\multirow{2}{*}{\textbf{Works}} & \multicolumn{3}{|c|}{\textbf{O - Sensing/Intuition (S/N)}} & \multicolumn{3}{|c|}{\textbf{C - Judging/Perceiving (J/P)}} & \multicolumn{3}{|c|}{\textbf{E - Extra/Introversion (E/I)}} & \multicolumn{3}{|c|}{\textbf{A - Sensing/Intuition (S/N)}} \\
\cline{2-13}
&      BA &      F1 &      RA &      BA &      F1 &      RA &      BA &      F1 &      RA &      BA &      F1 &      RA \\
\hline
 TAE \cite{Hu_He_Wang_Zhao_Shao_Nie_2024}& - & 0.8117 & - & - & 0.7020 & - & - & 0.7090 & - & - & 0.6621 & -\\
\hline
BERT MLP \cite{Mehta2020} & - & - & 0.6840 & - & - & 0.6440 & - & - & 0.7830 & - & - & 0.7440\\
\hline
MD-Transformer \cite{Yang_Quan_Yang_Yu_2021} & - & 0.6910 & - & - & 0.7919 & - & - & 0.6608 & - & - & 0.7919 & -\\
\hline
DGCN \cite{yang2023orders} & - & 0.6719 & - & - & 0.6816 & - & - & 0.6952 & - & - & 0.8053 & -\\
\hline
TrigNet \cite{yang-etal-2021-psycholinguistic} & - & 0.6717 & - & - & 0.6769 & - & - & 0.6954 & - & - & 0.7906 & -\\
\hline
CoT with Emotion \cite{li2024eerpdleveragingemotionemotion} & - & 0.9059 & \textbf{0.9101} & - & \textbf{0.8212*} & 0.8134 & - & \textbf{0.8663*} & 0.8710 & - & 0.8915 & 0.8917\\
\hline
\textbf{PICEPR (Contents Pipeline)} & \textbf{0.8858*}& \textbf{0.9569*} & \textbf{0.9268*} & \textbf{0.8158} & 0.7779 & \textbf{0.8190} & \textbf{0.8579} & 0.7596 & \textbf{0.8807} & \textbf{0.9039} & \textbf{0.9091} & \textbf{0.9032}\\
\hline
\textbf{PICEPR (Embeddings Pipeline)} &\textbf{0.8731} & \textbf{0.9273} & 0.8807 & \textbf{0.8563*} & \textbf{0.8254} & \textbf{0.8576*} & \textbf{0.8875*} & \textbf{0.7817} & \textbf{0.8847*} & \textbf{0.9276*} & \textbf{0.9333*} & \textbf{0.9280*}\\
\hline

\end{tabular}
\label{tab:kaggle-performance}
\begin{flushleft}
\footnotesize{Note: The \textbf{bold} text indicates the second-highest performance, while the \textbf{bold} text with an asterisk (\textbf{*}) represents the highest performance among all other algorithms. As this is an imbalanced dataset, BA metrics are used to reflect overall performance as they account for the minority class. 
The F1 score may not be as effective for evaluation since it does not consider true negatives (particularly for the `C' and `E' dimensions, where TN is larger than TP).}
\end{flushleft}
\end{table*}

\begin{figure}[ht]
    \centering
    \includegraphics[width=0.48\textwidth]{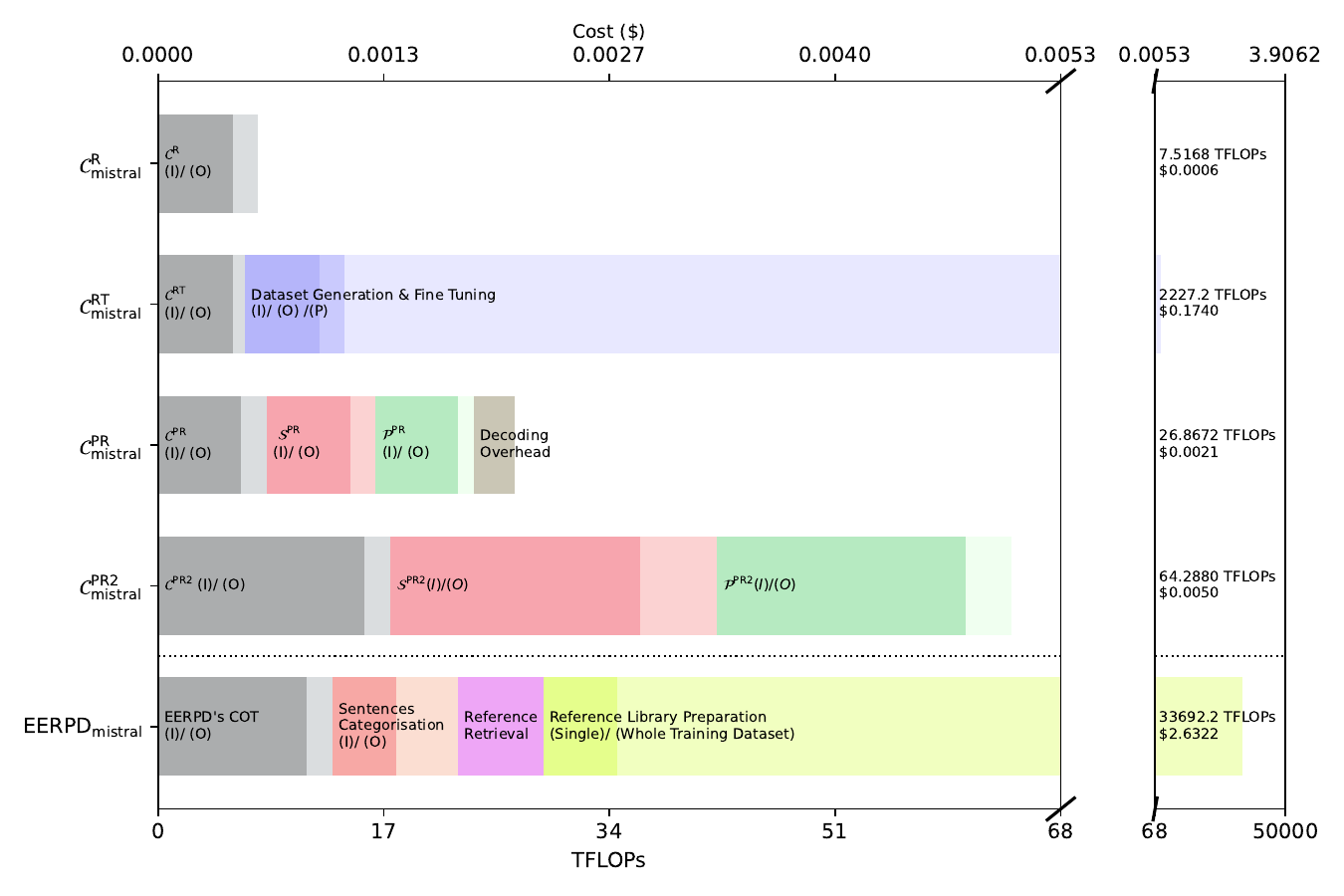}
 \caption{Cost analysis of $\mathcal{C}^{\text{R}}_{\text{mistral}}$, $\mathcal{C}^{\text{RT}}_{\text{mistral}}$, $\mathcal{C}^{\text{PR}}_{\text{mistral}}$, $\mathcal{C}^{\text{PR2}}_{\text{mistral}}$, and a reproduced $\text{EERPD}_{\text{mistral}}$ using the \textit{mistral} LLM as the backbone. (I) indicates input tokens, (O) indicates output tokens, and (P) stands for processing. In this analysis, we assume that the cost of input and output tokens is the same as stated in the \textit{mistral} documentation \cite{mistral_7b_2024}. However, to ensure a fair comparison, we include a decoding overhead applicable to other closed-source models such as \cite{gpt4o_2024}. This overhead is applied only to our $\mathcal{C}^{\text{PR}}_{\text{mistral}}$ to avoid underestimating its cost. For fine-tuning scenarios, we assume only one sample is involved in the process. For few-shot learning, we assume two samples are needed to satisfy the setting. For EERPD, since its performance depends on vectorizing the entire training dataset, the cost accounts for the whole dataset (as this contributes to its performance); however, we also visualize the case for only a single dataset using a darker yellowish-green color. Nonetheless, regardless of the setting, EERPD remains the highest-cost approach among all algorithms.}

    \label{fig:cost-analysis}
\end{figure}

To perform a cost analysis for the \textbf{Contents Pipeline}, we adopt the open-source \textit{mistral} model as our foundational backbone\cite{mistral_7b_2024}. This model offers a transparent computing platform, allowing us to estimate the floating-point operations (FLOPs) required for inference, as well as to reference the associated costs per input and output token \cite{openaipricing,DBLP:journals/corr/abs-2102-01293, mistral_7b_2024}. Figure~\ref{fig:cost-analysis} presents a comparative analysis of our four proposed algorithms within the Contents Pipeline, alongside the EERPD method~\cite{li2024eerpdleveragingemotionemotion}, which employs a decoder-only large language model (LLM). The results demonstrate that our proposed PICEPR algorithm—featuring a structured prompting mechanism—is both cost-efficient and effective for personality recognition tasks. Specifically, PICEPR incurs only twice the computational cost of the standard Chain-of-Thought (CoT) method while achieving more than a 10\% improvement in performance. Moreover, it outperforms the EERPD method by over 5\%, despite the latter requiring approximately 1200 times the computational resources. On the other hand, for the \textbf{Embeddings Pipeline}, which requires a training process, the data augmentation procedure undoubtedly increases computational demands. However, this step yields higher-quality data, thereby enhancing generalisation capabilities. These results further substantiate the effectiveness of our modularisation PICEPR approach in accurately understanding user personality.

\subsection{Limitation}

Despite the improvements achieved by PICEPR, we remain concerned about its generalizability across different personality labels. In line with prior research, we employed the widely used Essays and Kaggle datasets for our experiments. The Essays dataset was collected in a controlled experimental setting and annotated by psychologists \cite{Pennebaker1999}, whereas the Kaggle dataset consists of self-reported personality traits derived from social media platforms \cite{mitchell2017}. While PICEPR performs well on these datasets, we acknowledge their relatively small size and the inherent subjectivity of personality labels, which may introduce noise. 

Another limitation lies in content generation. Since LLMs are trained on vast amounts of internet data, their outputs may not strictly adhere to the principles of zero-shot learning. Some inference content may have been encountered during pretraining, meaning the model’s performance might partially depend on memorized patterns or knowledge rather than truly novel generation. This raises the possibility that the model's behavior is influenced by implicit pretraining biases, potentially resulting in “weight overwriting,” where certain content types dominate during training. 

Several researchers have expressed concerns regarding the inherent biases present in large language models (LLMs) \cite{Raza2025,Caliskan2017}. These models can perpetuate traditional stereotypes, particularly when applied to tasks such as personality recognition. For example, Xu, et al demonstrated that machine learning models trained on large corpora of text can exhibit human-like biases, including those related to race, gender, and social roles \cite{Xu2025}. In the context of personality recognition, our daily activities may correlate with certain personality traits, potentially leading LLMs to make biased classifications. However, in our experiment, the Essays dataset is annotated by trained professionals, which yielded also reliable and consistent results. This suggests that the risk of bias was mitigated to a significant extent. Therefore, although the risk of bias in personality inference remains a valid concern, our findings suggest that the use of carefully curated and professionally annotated data provides a robust basis for accurate classification. Put differently, behavioral cues embedded in everyday activities, even when individually subtle, can collectively serve as informative features for personality inference. This approach is similar to psychological assessment practices, where the accumulation of small behavioral indicators over time supports a more accurate identification of underlying personality traits \cite{Caliskan2017}.

These highlights the need to evaluate the approach on larger and more diverse datasets as they become available. Due to collecting a new and large dataset is currently not feasible due to GPU resources and psychologist involvement for annoation for new dataset, a future work could be rerun the experiments on a larger dataset.

\section{Conclusion}

The proposed PICEPR achieves a new state-of-the-art model for personality recognition by 5-15\% in the Essays and Kaggle dataset to the regular approaches, and we can answer the RQs:
\begin{itemize}
    
    \item \textbf{RQ1:} PICEPR effectively addresses both hallucination, where the model generates irrelevant outputs, and also the lost-in-the-middle phenomenon, where critical information is misrepresented or lost during the decoding process. A modularised approach could help the LLM focus on psychological features rather than the main task of classifying personality. By modularised tasks, the input token length for the classification model is reduced, leading to more focused and efficient processing. While this modular design roughly doubles the computational cost, our experiments demonstrate that the PICEPR approach remains effective—even when using less advanced LLMs—by producing more accurate and reliable outputs despite the underlying model limitations.

    \item \textbf{RQ2:} 
    Although closed-source \textit{gpt4o} models typically provide superior performance, approaches like PICEPR illustrate that large language models (LLMs) can be highly effective as intermediate modules for analyzing textual content, making inferences, and even generating augmented training data to enhance personality classification. This study contributes valuable insights into how LLMs can be leveraged for psychological tasks, demonstrating their potential beyond traditional NLP applications. In particular, the proposed PICEPR algorithm effectively addresses the class imbalance issue inherent in personality datasets. This imbalance naturally arises due to the uneven distribution of personality traits in social media data, such as the Kaggle dataset where label distributions are skewed. This approach leads to more robust personality recognition, as LLMs contribute richer personality insights that enhance overall predictive power.

    \item \textbf{RQ3:} 
    The experiments show that fine-tuning decoder-only models did not yield significant improvements, regardless of the underlying LLM architecture. However, this limitation can be addressed through an effective prompting framework—specifically, the proposed PICEPR (Contents Pipeline)—which leverages the generative capabilities of decoder-only models to extract personality cues for accurate classification. In contrast, fine-tuning encoder-only LLMs on limited or low-quality datasets often results in poor generalisation. To overcome this, the proposed PICEPR (Embeddings Pipeline) enables the generation of high-quality datasets that significantly enhance generalisation performance. PICEPR demonstrates its ability to extract personality insights through both the Contents and Embeddings pipelines, each offering distinct strengths and trade-offs. The Contents Pipeline allows direct utilization of decoder-only LLMs in a zero-shot manner without requiring any additional training. Meanwhile, the Embeddings Pipeline transfers the reasoning ability of decoder-only models to encoder-only models via dataset augmentation, followed by fine-tuning. While this approach requires additional training effort, it results in significantly faster and more computationally efficient inference.

\end{itemize}

Future work could focus on investigating the Embeddings Pipeline, as the augmentation techniques improve model bias but do not sufficiently resolve the imbalance problem. Additionally, research could explore methods to ensure that decoder models generate outputs with reduced computational resources while minimizing hallucination.

\section*{Acknowledgment}
This research was funded by the Universiti Tunku Abdul Rahman Research Fund (IPSR/RMC/UTARRF/2021-C1/K03). The authors also appreciate the support of Grid5000 for providing the computational resources used in this study.   
We also appreciate National Yang Ming Chiao Tung University for providing the Research Scholarship to establish professional guidance related to psychology. We appreciate the Embassy of France to Malaysia for the Doctoral Research Mobility Grant, which facilitated the research collaboration between Universiti Tunku Abdul Rahman, and Université Sorbonne Paris Nord.

\bibliographystyle{IEEEtran}
\bibliography{references}

\end{document}